\documentclass{article}

\PassOptionsToPackage{numbers, compress}{natbib}
\usepackage[preprint]{neurips_2026}

\usepackage{hyperref}       

\hypersetup{
    colorlinks=true,
    linkcolor=red,
    citecolor=blue,
    urlcolor=blue,
}

\setcitestyle{numbers,square,comma}

\usepackage{xurl}
\usepackage{url}            
\usepackage{booktabs}       
\usepackage{amsfonts}       
\usepackage{nicefrac}       
\usepackage{microtype}      
\usepackage{xcolor}         
\usepackage{multirow}       
\usepackage{algorithm}      
\usepackage{algorithmic}    
\usepackage{amsmath}        
\usepackage{graphicx}
\usepackage{tcolorbox}
\usepackage{tabularx} 
\usepackage{subcaption}
\usepackage{placeins}       

\title{KernelGenBench: Can LLMs and Agents Write Efficient Kernels Across Operator Sources and Hardware Platforms?}

\author{%
  \bfseries Peiyu Zang\textsuperscript{1,5}, Jian Tao\textsuperscript{5}, Jialing Zhang\textsuperscript{2,5}, Yichen Yuan\textsuperscript{3,5}, \\
  \bfseries Wentao Zhang\textsuperscript{4}, Guang Liu\textsuperscript{5}\thanks{Corresponding author.}, Yonghua Lin\textsuperscript{5} \\[0.2em]
  \normalfont \textsuperscript{1}Beijing Normal University, \textsuperscript{2}Beijing Jiaotong University, \\
  \normalfont \textsuperscript{3}Institute of Automation, CAS, \textsuperscript{4}Peking University, \textsuperscript{5}BAAI \\[0.2em]
  \texttt{{202421130105@mail.bnu.edu.cn, liuguang@baai.ac.cn}}
}

\begin{document}

\maketitle

\begin{abstract}
Modern AI systems depend on specialized accelerator kernels, whose development is complicated by increasingly diverse operators and hardware. LLMs and agentic systems promise to automate this work, but existing evaluations do not show whether their performance transfers across operator sources and hardware platforms, or what such transfer costs. We present \textbf{KernelGenBench}\footnote{Code and data are publicly available at \url{https://github.com/flagos-ai/KernelGenBench}.}, the first unified multi-source $\times$ multi-chip infrastructure for evaluating LLM- and agent-generated Triton kernels. With a common Triton target spanning six hardware platforms, it provides the broadest cross-vendor hardware coverage among existing kernel-generation benchmarks. We report two controlled analytical views: \textbf{KernelGenBench-MS (Multi-Source)} covers 210 operators from PyTorch ATen, production vLLM operators, and proprietary cuBLAS routines, while \textbf{KernelGenBench-MC (Multi-Chip)} evaluates a semantically stable 110-operator subset across six hardware platforms. Our evaluation consumed over 15 billion tokens. More interactive agentic scaffolds achieved higher correctness, but no method dominated across sources and platforms: vLLM posed the strongest correctness challenge, cuBLAS set the highest performance ceiling, and AutoKernel accuracy fell from 87\% on NVIDIA to 25\% on Iluvatar CoreX. These improvements were costly: specialized agents averaged 4.99 million tokens per successful operator, rising to 6.25 million for CUDA Optimized Skill. The results establish operator source, hardware platform, and agentic scaffold as distinct dimensions of kernel-generation capability, and show that success in a familiar source--hardware setting is not a reliable proxy for deployment readiness.
\end{abstract}

\section{Introduction}
\label{sec:intro}

As AI models grow, the efficient use of compute infrastructure has become a major constraint on their development and deployment. This infrastructure is also becoming more heterogeneous: GPUs, TPUs, and other accelerators collectively provide rapidly growing compute capacity and memory bandwidth~\cite{stanford2026aiindex,epoch2026hbmshipped}. Provisioning more compute, however, does not by itself improve application performance. Kernel implementations determine how tensor operators use hardware parallelism and memory hierarchies. Writing efficient kernels still requires expertise in low-level programming, hardware architecture, and performance optimization. Recent work uses LLMs and agents to automate parts of this process through iterative generation, profiling, and compiler-guided optimization~\cite{liu2026dr,wang2025geak,yang2026atrex,shui2026harness,ye2026cake}.

\textbf{Can kernel-generation systems transfer across operator sources and hardware platforms?} Recent model architectures make this question more pressing. DeepSeek V4 introduces compressed sparse attention, which requires model-specific kernels beyond standard framework operators~\cite{deepseek2026v4}. Production inference frameworks such as vLLM and SGLang likewise rely on specialized CUDA kernels in performance-critical paths~\cite{kwon2023efficient,zheng2024sglang}. These models must also be trained and served on heterogeneous accelerators. A PyTorch-only evaluation therefore covers only part of the problem: kernel-generation systems must implement framework-native semantics and production CUDA operators, compete with highly optimized libraries, and transfer across hardware platforms. Recent benchmarks extend task taxonomies, agentic workflows, verification, production realism, and hardware-specific evaluation~\cite{wang2026kernelbenchx,guan2026tritongym,zhang2026kernelbenchverified,yang2026atrex,huang2026realistictritonbench,zhang2026ptxbench}. Existing benchmarks have broadened either operator provenance or hardware coverage, but none unifies distinct operator-reference ecosystems with controlled evaluation across heterogeneous accelerator software stacks under a common kernel-generation and verification framework. Such unification requires more than adding device targets: it must integrate distinct reference implementations, compilers, runtimes, and backend APIs through a common task interface and an extensible platform-adapter layer. A kernel-generation system that performs well for the same operator on NVIDIA may behave quite differently on another hardware platform, where hardware architectures and software stacks differ substantially.

We introduce \textbf{KernelGenBench}, the first unified multi-source $\times$ multi-chip benchmark infrastructure. Its common Triton target spans six hardware platforms, providing the broadest cross-vendor hardware coverage among existing kernel-generation benchmarks. To distinguish source-conditioned behavior from hardware-stack transfer, we report two complementary controlled views. Triton's programmable block-level abstraction supports model-specific fusion, data movement, and scheduling, allowing generated kernels to be evaluated against framework-native operators, production CUDA kernels, and tuned libraries rather than only compositions of PyTorch APIs. Its compiler-based portability provides a common kernel language across hardware platforms and enables controlled tests of performance portability~\cite{tillet2019triton,ansel2024pytorch}. We instantiate these views as:
\begin{itemize}
    \item \textbf{KernelGenBench-MS (Multi-Source):} Evaluates kernel generation across 210 operators from PyTorch ATen~\cite{paszke2019pytorch}, vLLM~\cite{kwon2023efficient}, and cuBLAS.
    \item \textbf{KernelGenBench-MC (Multi-Chip):} Assesses performance portability across six hardware platforms, including NVIDIA, Moore Threads, Hygon, Iluvatar CoreX, and two anonymous platforms, using a 110-operator ATen subset.
\end{itemize}

\begin{figure*}[t]
\centering
\includegraphics[width=\textwidth]{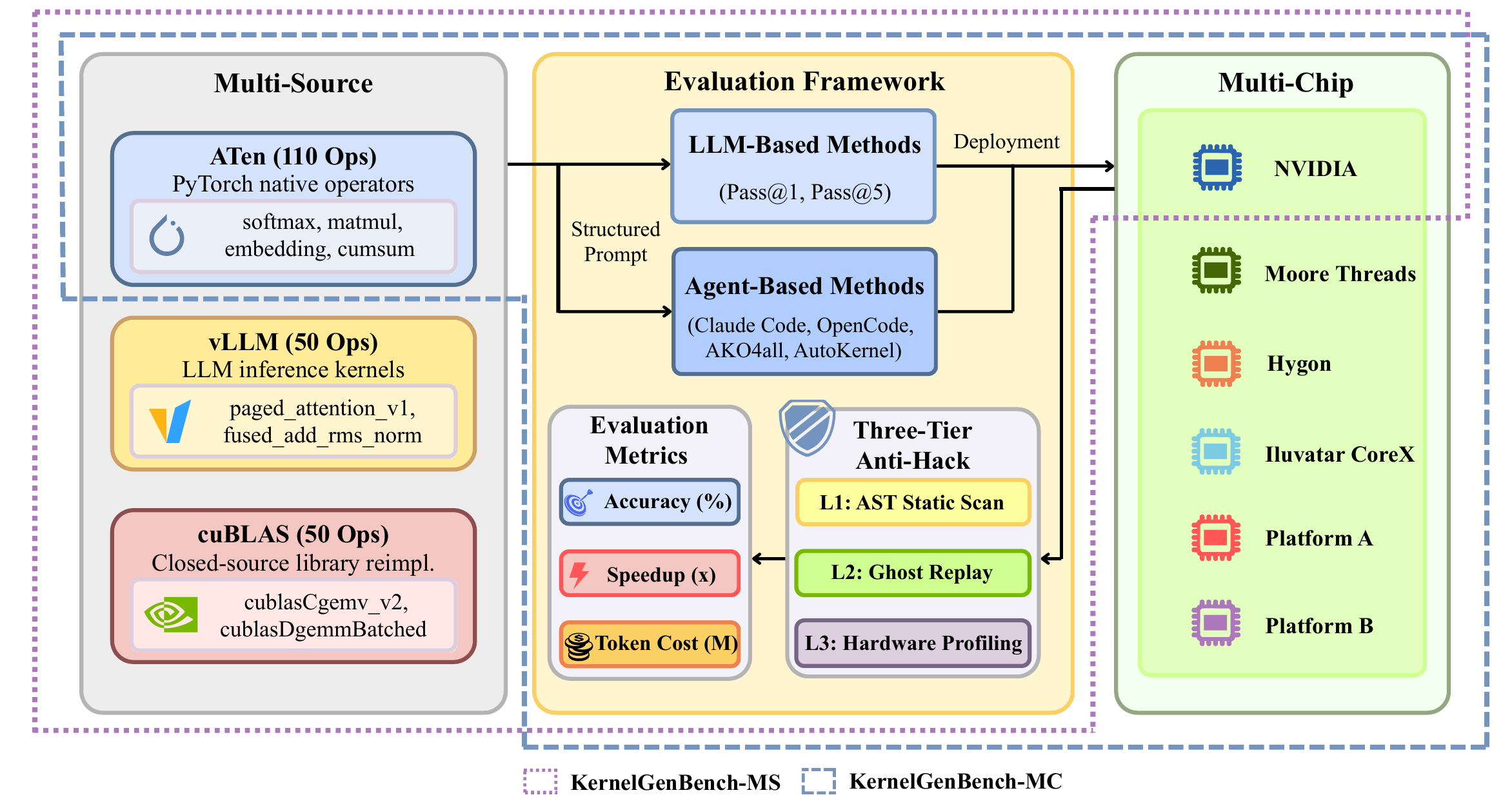}
\caption{Overview of KernelGenBench. The shared infrastructure supports multi-source $\times$ multi-chip evaluation; the purple and blue dashed boxes indicate the MS and MC controlled analytical views reported in this paper.}
\label{fig:overview}
\end{figure*}

For execution-grounded evaluation, we design a framework with a distributed sandbox infrastructure and a three-tier anti-hack mechanism that prevents benchmark evasion. We also introduce a multi-dimensional evaluation protocol that jointly measures functional correctness, execution speedup, and agentic cost efficiency.

Our evaluation consumed over 15 billion tokens and produced three main findings. (1) Agent-based methods consistently outperform pure LLM sampling. Across operator sources, ATen is the most tractable; vLLM is the hardest to generate correctly but offers substantial speedup potential; and cuBLAS is the strongest performance-ceiling stress test. (2) Performance varies substantially across hardware platforms. Kernel-specialized agents can degrade sharply outside NVIDIA: AutoKernel accuracy falls from 87\% on NVIDIA to 25\% on Iluvatar CoreX. (3) More interactive agentic scaffolds achieve higher correctness at high cost. Specialized methods average 4.99 million tokens per successful operator, rising to 6.25 million for CUDA Optimized Skill. KernelGenBench quantifies how these effects vary across operator sources and hardware platforms, together with their agentic cost.

Our contributions are threefold:
\begin{itemize}
    \item We build the first unified multi-source $\times$ multi-chip benchmark infrastructure, coupling three operator sources with platform-specific runtime and library adapters. We report MS across 210 operators on NVIDIA and controlled MC evaluation across six hardware platforms using a common 110-operator task set (Section~\ref{sec:suite}).
    \item We develop a common evaluation framework combining combinatorial correctness tests, performance measurement, cost accounting, and layered anti-hack verification.
    \item We evaluate LLM sampling and agent-based methods at scale and identify distinct source-level capability boundaries, cross-platform portability gaps, and the benefits and costs of iterative agentic workflows.
\end{itemize}

\section{Related Work}
\label{sec:related}

\subsection{LLM-Driven Kernel Generation}

Automated kernel generation combines model training with inference-time search~\cite{yu2026automatedkernelgenerationera}. Prior work explores supervised fine-tuning, reinforcement learning, execution feedback, evolutionary search, and runtime-guided refinement~\cite{kernelllm2025,baronio2025kevin,yang2026incoder,dai2026cuda,chen2026avo,hammond2025agentic}. More recent systems extend these ideas to AMD targets, architecture-specific PTX, profile-driven optimization, engineered harnesses, and agent-facing compiler abstractions~\cite{wang2025geak,zhang2026ptxbench,yang2026atrex,shui2026harness,ye2026cake}. These evaluations therefore characterize the complete model--agent--toolchain scaffold rather than the language model in isolation.

\subsection{Limitations of Existing Benchmarks}

KernelBench and TritonBench established execution-based evaluation for generated kernels, while subsequent benchmarks broaden task categories, agentic workflows, hidden verification, memory metrics, and cross-hardware analysis~\cite{ouyang2025kernelbench,li2025tritonbench,wang2026kernelbenchx,guan2026tritongym,zhang2026kernelbenchverified}. Production-oriented efforts further derive workloads from serving traces or real framework pull requests and evaluate richer execution contexts~\cite{saroufim2025backendbench,yang2026atrex,huang2026realistictritonbench}. These efforts diversify task categories, architectures, or production contexts; KernelGenBench instead holds a common API-replacement protocol fixed while comparing operator collections with distinct sources and reference implementations: ATen, vLLM, and cuBLAS. It additionally evaluates layered anti-hack enforcement and agentic cost.

Most evaluation nevertheless remains NVIDIA-centric or deliberately architecture-specific~\cite{xing2026flashinferbench,lin2026solexecbench,zhang2026ptxbench}. Atrex-Bench draws operators from multiple production serving stacks but presents them through a normalized PyTorch-reference interface~\cite{yang2026atrex}. MultiKernelBench spans NVIDIA GPUs, Huawei NPUs, and Google TPUs using separate platform-specific languages~\cite{wen2025multikernelbench}; KernelGenBench instead uses one Triton target across six hardware platforms. Other benchmarks extend evaluation toward AMD or cross-hardware settings~\cite{wang2025geak,yang2026atrex,wang2026kernelbenchx}. Applying one verification pipeline and one 110-operator task set across these platforms enables correctness, speedup, and cost comparisons under a common protocol. Direct CUDA or CUTLASS generation remains valuable for NVIDIA peak performance, but cannot serve as the shared target for this multi-chip axis; we therefore use Triton and empirically evaluate, rather than assume, portability.

\section{KernelGenBench}
\label{sec:suite}
KernelGenBench is a unified multi-source $\times$ multi-chip benchmark infrastructure. Figure~\ref{fig:overview} illustrates the two controlled views: KernelGenBench-MS and KernelGenBench-MC.

\subsection{KernelGenBench-MS}
\label{subsec:taxonomy}
We define multi-source evaluation as evaluation across operator collections that differ in source, abstraction level, and reference implementation. KernelGenBench-MS is anchored on a stable NVIDIA baseline and curates 210 operators from three complementary sources: framework-native operators (PyTorch ATen), production inference-specific CUDA operators (vLLM), and routines from a vendor-optimized proprietary library (cuBLAS). This progression separates three capabilities that a PyTorch-only suite conflates: reproducing standardized framework semantics, reconstructing specialized operators from production-facing specifications, and approaching the performance ceiling of expert-optimized libraries. We provide detailed API-level context, but none of the three subsets exposes the underlying C++/CUDA production kernel to the model. Generated Triton kernels must exactly match the specifications as drop-in replacements~\cite{flaggems2024}. Source-specific prompt context is detailed in Appendix~\ref{app:prompt_construction}; full operator lists are in Appendix~\ref{app:full_op_list}.

\textbf{Source 1: PyTorch ATen Operators (110 problems).} From the 900+ \texttt{torch.ops.aten} APIs, we select the 50 most frequent operators in training traces from 2{,}907 open-source models and uniformly sample 60 long-tail operators. Each operator API is an independent problem (e.g., \texttt{torch.abs} and \texttt{torch.abs\_} are two separate problems), requiring the model to implement all overload variants. Prompts use the official \texttt{FunctionSchema} and docstrings; the reference calls the corresponding native ATen API.

\textbf{Source 2: vLLM Operators (50 problems).} This subset represents the specialized kernels that make modern inference runtimes efficient and that must be reimplemented when their workloads move beyond their original hardware platform. We select vLLM operators with standalone CUDA implementations, including attention mechanisms and KV-cache management.\footnote{All vLLM evaluations use v0.13.0 in this paper.} The prompt includes the benchmark's Python baseline wrapper, including its signature, argument order, in-place behavior, and custom-operator invocation, but not the CUDA/C++ implementation behind that operator. It therefore tests whether a system can recover production-specific semantics and produce a drop-in Triton implementation without copying the reference kernel.

\textbf{Source 3: cuBLAS Operators (50 problems).} Correctness on framework or runtime operators does not establish that a generated kernel can approach expert-tuned performance. We therefore designate the cuBLAS closed-source dynamic library as a vendor-optimized performance-ceiling baseline.\footnote{All cuBLAS evaluations use v12.4 in this paper.} The baseline loads \texttt{libcublas.so} directly via \texttt{ctypes.cdll}, bypassing all high-level wrappers. To construct a balanced problem set, we first identify the 10 most frequently invoked cuBLAS routines via profiling traces and expand them across precision and batching modes (Appendix~\ref{app:cublas_gemm}). To avoid over-indexing on one operator family, we also sample functionally diverse BLAS routines, requiring models to match proprietary performance without external libraries.

\subsection{KernelGenBench-MC}
\label{subsec:multichip}
KernelGenBench provides a multi-source $\times$ multi-chip evaluation infrastructure: a common pipeline couples the three operator sources with platform-specific runtime and library adapters. For controlled analysis, we report two views of this infrastructure. MS fixes NVIDIA and varies source over all 210 operators, characterizing source-conditioned generation and optimization behavior. MC fixes the semantically stable 110-operator ATen suite and varies six hardware stacks, characterizing end-to-end hardware-stack transfer. We do not pool the full cross-product into one main-table ranking: production-runtime ports and vendor-library counterparts vary in coverage, plugin maturity, implementation path, and performance ceiling. The same nominal operator can invoke a native CUDA custom op, a vendor adaptation, a compatibility layer, or no local implementation. Pooling these cases would mix source difficulty with stack maturity. The cross-vendor reference-adapter layer underlying these configurations is summarized in Table~\ref{tab:cross_vendor_adapters}. The MS and MC views therefore analyze source-conditioned and hardware-stack variation separately.

The framework targets six hardware platforms, including NVIDIA, Moore Threads, Hygon, Iluvatar CoreX, and two anonymous platforms. Two platforms remain de-identified under commercial confidentiality requirements. KernelGenBench-MC uses Triton as the common target language because direct CUDA/CUTLASS generation is NVIDIA-only. Its single pipeline detects the hardware, injects platform-specific constraints, and applies adaptive tolerances and anti-hack configurations (Appendix~\ref{app:sandbox_chip}).

\subsection{Evaluation Framework}
\label{subsec:pipeline_and_antihack}

All KernelGenBench configurations use a shared evaluation infrastructure that executes the following workflow. The distributed sandbox infrastructure is detailed in Appendix~\ref{app:sandbox_infra}.

\textbf{Kernel Generation.}
Under the same operator specification, test suite, applicable anti-hack checks, and 30-minute task-level wall-clock budget, we evaluate two paradigms: (1) \emph{LLM Sampling Methods}, comprising direct zero-shot and multi-sample generation (Pass@1, Pass@5)~\cite{chen2025pass} without execution feedback (pseudocode in Appendix~\ref{app:pipeline_algorithms}); and (2) \emph{Agent-Based Methods}, comprising autonomous coding agents with execution feedback loops, including vanilla agentic frameworks and kernel-specialized agents. Because prompts, skills, profilers, controllers, and tool access shape agentic kernel generation~\cite{guan2026tritongym,shui2026harness}, we standardize external task conditions and final verification while retaining these method-specific mechanisms (Appendix~\ref{app:pipeline_algorithms}).

\textbf{Verification.}
Generated kernels serve as input to the unified \texttt{verify} engine. Unlike fixed single-shape evaluation~\cite{ouyang2025kernelbench}, we construct a combinatorial test suite of $k_i$ distinct test cases per operator by taking the Cartesian product of core semantic parameters (e.g., \emph{dim}, \emph{transpose}) alongside shape, data type, and memory layout (details in Appendix~\ref{app:sandbox_stress}). This design is complementary to hidden-distribution verification~\cite{zhang2026kernelbenchverified} and native-framework end-to-end testing~\cite{huang2026realistictritonbench}. Kernels are mounted directly into PyTorch's dispatch tree via \texttt{torch.library} to test drop-in replaceability. A kernel is considered correct only when it passes numerical validation across \emph{all} $k_i$ test cases.

\textbf{Anti-Hack Architecture.}
To guard against \textbf{benchmark hacking}~\cite{tritonrl2025,zhang2026kernelbenchverified} (e.g., bypassing Triton compilation by calling pre-compiled backend APIs or exploiting a narrow test distribution), we deploy a three-tier interception mechanism:
\begin{itemize}
    \item \emph{L1: AST Static Scan}: Parses the generated abstract syntax tree to block blacklisted calls (e.g., \texttt{torch.ops.aten.*}, \texttt{import vllm}, and \texttt{ctypes}) and dynamic injection bypasses.
    \item \emph{L2: Ghost Replay}: First executes the kernel normally to capture outputs, then replaces the \texttt{@triton.jit}-decorated function with a no-op in memory and replays; identical outputs indicate that the Triton kernel was not causally used, triggering a cheating flag.
    \item \emph{L3: Hardware Profiling}: Uses \texttt{torch.profiler} to ensure Triton-specific signatures exist in low-level trace logs. This layer operates exclusively on NVIDIA hardware; heterogeneous platforms rely on the first two layers due to the absence of equivalent profiling tools.
\end{itemize}

These checks are empirically necessary: during evaluation, they intercepted 13 Pass@1 and 30 Pass@5 shortcut submissions, including direct target-API reuse and hybrid framework fallback; Appendix~\ref{app:generation_behaviors} reports representative cases and interception breakdowns.

\textbf{Evaluation Metrics.}
We evaluate correctness, performance, and cost efficiency. Clean pass rate counts an operator as successful only when all $k_i$ test cases pass numerical validation and all anti-hack layers available on that platform. For performance, we compute a two-level geometric mean~\cite{spec2017cpu, fleming1986not}: first across the $k_i$ test cases to obtain operator-level speedup $S_i$, then across the operators successfully solved by that method; its definition and rationale are detailed in Appendix~\ref{app:speedup_metrics}. Global speedup is therefore a successful-set statistic, not a matched-set ranking across methods. Where complete, comparable accounting is available, we report total tokens, tokens per success (total tokens divided by passed operators), and total wall time (cumulative per-operator solving duration, parallelism-agnostic).
\begin{table}[!t]
\centering
\setlength{\tabcolsep}{4pt}
\caption{NVIDIA A100 evaluation across 210 operators from three sources. Spd is computed on each method's own correctly solved operator set and is not a matched-set cross-method ranking.}
\label{tab:nvidia_main_ops}
\resizebox{\textwidth}{!}{%
\begin{tabular}{l | cc | cc | cc | cc}
\toprule
\multirow{2}{*}{\textbf{Method \& Setup}} & \multicolumn{2}{c|}{\textbf{Overall (210)}} & \multicolumn{2}{c|}{\textbf{ATen (110)}} & \multicolumn{2}{c|}{\textbf{vLLM (50)}} & \multicolumn{2}{c}{\textbf{cuBLAS (50)}} \\
\cmidrule(lr){2-3} \cmidrule(lr){4-5} \cmidrule(lr){6-7} \cmidrule(lr){8-9}
& Acc (\%) & Spd ($\times$) & Acc (\%) & Spd ($\times$) & Acc (\%) & Spd ($\times$) & Acc (\%) & Spd ($\times$) \\
\midrule
\multicolumn{9}{l}{\textbf{LLM Sampling Methods}} \\
\midrule
Pass@1 (Opus-4.6)       & 41 & 0.70 & 39 & 0.90 & 20 & 0.76 & 68 & 0.49 \\
Pass@1 (GLM-5.0)        & 21 & 0.68 & 21 & 0.49 & 24 & 1.24 & 20 & 0.73 \\
Pass@1 (Qwen3.5-27b)    & 7  & 0.85 & 8  & 0.83 & 2  & \textbf{2.05} & 8  & 0.71 \\
Pass@1 (MiniMax M-2.5)  & 2  & 0.88 & 4  & 0.88 & 0  & -- & 0  & -- \\
\midrule
Pass@5 (Opus-4.6)       & 57 & 0.68 & 62 & 0.79 & 28 & 0.71 & 74 & 0.49 \\
Pass@5 (GLM-5.0)        & 36 & 0.77 & 45 & 0.64 & 32 & 1.28 & 20 & \underline{0.76} \\
Pass@5 (Qwen3.5-27b)    & 11 & \underline{1.01} & 13 & \textbf{1.04} & 12 & 0.70 & 8  & 0.68 \\
Pass@5 (MiniMax M-2.5)  & 16 & 0.69 & 21 & 0.76 & 18 & 1.27 & 2  & 0.46 \\
\midrule
\multicolumn{9}{l}{\textbf{Vanilla Agentic Frameworks}} \\
\midrule
Claude Code (Opus-4.6)     & 87 & 0.78 & 92 & 0.86 & 68 & 1.02 & 94 & 0.51 \\
Claude Code (GLM-5.0)      & 67 & 0.83 & 72 & 0.88 & 52 & 1.23 & 72 & 0.53 \\
Claude Code (Qwen3.5-27b)  & 62 & 0.70 & 80 & 0.68 & 38 & 1.17 & 48 & 0.50 \\
Claude Code (MiniMax M-2.5)& 49 & 0.69 & 69 & 0.78 & 26 & 0.46 & 26 & 0.58 \\
\midrule
OpenCode (Opus-4.6)        & 84 & 0.80 & 90 & 0.84 & 58 & 1.35 & 96 & 0.53 \\
OpenCode (GLM-5.0)         & 70 & 0.74 & 87 & 0.75 & 46 & 1.23 & 54 & 0.46 \\
OpenCode (Qwen3.5-27b)     & 53 & 0.78 & 58 & 0.75 & 44 & 1.31 & 52 & 0.58 \\
OpenCode (MiniMax M-2.5)   & 41 & 0.62 & 50 & 0.77 & 26 & 0.44 & 36 & 0.42 \\
\midrule
\multicolumn{9}{l}{\textbf{Kernel-Specialized Agents}} \\
\midrule
AKO4all (Opus-4.6)              & 83 & 0.97 & 91 & \underline{1.00} & 64 & 1.62 & 84 & 0.61 \\
CUDA Opt. Skill (MiniMax M-2.5) & 45 & 0.80 & 63 & 0.81 & 24 & 0.92 & 28 & 0.45 \\
\midrule
AutoKernel (GLM-5.0)            & 71 & 0.99 & 87 & \underline{1.00} & 42 & 1.40 & 66 & 0.75 \\
AutoKernel (Qwen3.5-27b)        & 47 & \textbf{1.02} & 69 & \underline{1.00} & 16 & \underline{1.63} & 30 & \textbf{0.80} \\
AutoKernel (MiniMax M-2.5)      & 43 & 0.89 & 66 & 0.87 & 20 & 1.52 & 16 & 0.46 \\
\bottomrule
\end{tabular}%
}
\end{table}

\section{Experiments and Evaluation}
\label{sec:experiments}
\setcounter{topnumber}{1}
\setcounter{bottomnumber}{0}

We use KernelGenBench to study three questions that single-source, single-platform accuracy cannot answer: whether capability transfers across operator source, whether it transfers across hardware platforms, and how agentic scaffolds alter the accuracy--performance--cost trade-off.

\subsection{Experimental Setup}
\label{subsec:setup}

Following the benchmark design in Section~\ref{sec:suite}, we report the MS view on NVIDIA across 210 operators and the MC view across six hardware platforms. Main-table Pass@1 configurations use \texttt{temperature}=0 and Pass@5 uses \texttt{temperature}=0.8, with \texttt{max\_tokens}=16384 and a unified 30-minute wall-clock timeout per operator task. The full study consumed over 15 billion tokens and approximately \$20{,}000 in API expenditure. Most agent results are therefore single-trajectory measurements, and differences below 5~pp are interpreted cautiously. Kernel performance is measured separately with warm-up and repeated timing runs. Timing and repeated-generation protocols are detailed in Appendices~\ref{app:repeated_run_robustness} and~\ref{app:detailed_limitations}.

\subsection{KernelGenBench-MS: Multi-Source Evaluation}
\label{subsec:nvidia_analysis}

Table~\ref{tab:nvidia_main_ops} presents the results; additional NVIDIA results are reported in Appendix~\ref{app:legacy_baselines}.

\textbf{Finding 1: Operator source determines which capability boundary is exposed.}
Claude Code (Opus-4.6) achieves the highest overall accuracy at 87\%, while AutoKernel~\cite{jaber2026autokernel} (Qwen3.5-27b) reports the highest successful-set overall speedup at 1.02$\times$. Claude Code and AKO4all~\cite{ako4all2026} (Opus-4.6) obtain 87\% and 83\% accuracy, respectively, while their successful-set speedups are 0.78$\times$ and 0.97$\times$. The 4~pp accuracy gap is descriptive under single-trajectory evaluation, not a statistically established ordering. Because speedups use different successful sets, they should not alone rank methods. Detailed distributions are in Appendix~\ref{app:fastp_results}; additional model-specific generation behaviors are analyzed in Appendix~\ref{app:generation_behaviors}.

\textbf{Correctness difficulty and optimization difficulty do not share a single ordering.}
The three sources expose complementary difficulty profiles: ATen is comparatively tractable, vLLM stresses semantic reconstruction while leaving greater optimization headroom, and cuBLAS permits high correctness but sets a much stronger performance ceiling. Section~\ref{subsec:source_profiles} analyzes these source-conditioned differences across configurations and operators.

\subsection{KernelGenBench-MC: Cross-Platform Evaluation}
\label{subsec:cross_platform_analysis}

We extend the evaluation of the 110 ATen operators across six hardware platforms. Table~\ref{tab:cross_platform_main} presents the full results.

\begin{table}[!t]
\centering
\setlength{\tabcolsep}{3.5pt}
\caption{Cross-platform evaluation on 110 ATen operators. Spd is computed on each method's own correctly solved operator set; it is not a matched-set cross-method ranking.}
\label{tab:cross_platform_main}
\resizebox{\textwidth}{!}{%
\begin{tabular}{l | cc | cc | cc | cc | cc | cc}
\toprule
\multirow{2}{*}{\textbf{Method \& Setup}} & \multicolumn{2}{c|}{\textbf{NVIDIA}} & \multicolumn{2}{c|}{\textbf{Moore Threads}} & \multicolumn{2}{c|}{\textbf{Hygon}} & \multicolumn{2}{c|}{\textbf{Iluvatar CoreX}} & \multicolumn{2}{c|}{\textbf{Platform A}} & \multicolumn{2}{c}{\textbf{Platform B}} \\
\cmidrule(lr){2-3} \cmidrule(lr){4-5} \cmidrule(lr){6-7} \cmidrule(lr){8-9} \cmidrule(lr){10-11} \cmidrule(lr){12-13}
 & Acc (\%) & Spd ($\times$) & Acc (\%) & Spd ($\times$) & Acc (\%) & Spd ($\times$) & Acc (\%) & Spd ($\times$) & Acc (\%) & Spd ($\times$) & Acc (\%) & Spd ($\times$) \\
\midrule
\multicolumn{13}{l}{\textbf{LLM Sampling Methods}} \\
\midrule
Pass@1 (Opus-4.6) & 39 & 0.90 & 44 & 0.69 & 37 & 0.98 & 38 & 0.88 & 46 & 0.19 & 38 & 0.89 \\
Pass@1 (Qwen3.5-27b) & 8 & 0.83 & 3 & 1.02 & 7 & 0.90 & 10 & 1.03 & 9 & 0.09 & 7 & 0.98 \\
Pass@1 (MiniMax M-2.5) & 4 & 0.88 & 6 & \underline{1.33} & 4 & \underline{1.05} & 4 & \textbf{1.15} & 4 & 0.25 & 5 & 0.77 \\
\midrule
Pass@5 (Opus-4.6) & 62 & 0.79 & 60 & 0.74 & 54 & 0.92 & 57 & 0.83 & 63 & 0.15 & 65 & 0.68 \\
Pass@5 (Qwen3.5-27b) & 13 & \textbf{1.04} & 11 & 1.10 & 15 & 0.72 & 17 & 1.02 & 16 & 0.18 & 10 & 0.99 \\
Pass@5 (MiniMax M-2.5) & 21 & 0.76 & 15 & 0.53 & 12 & \underline{1.05} & 9 & 0.76 & 17 & 0.20 & 8 & 0.33 \\
\midrule
\multicolumn{13}{l}{\textbf{Vanilla Agentic Frameworks}} \\
\midrule
Claude Code (Opus-4.6) & 92 & 0.86 & 93 & 0.80 & 88 & 0.87 & 83 & 0.83 & 89 & 0.18 & 96 & 0.89 \\
Claude Code (GLM-5.0) & 72 & 0.88 & 65 & 0.96 & 65 & 0.81 & 37 & 0.77 & 65 & 0.16 & 59 & 0.90 \\
Claude Code (Qwen3.5-27b) & 80 & 0.68 & 75 & 0.61 & 75 & 0.85 & 23 & 0.81 & 78 & 0.25 & 82 & 0.77 \\
Claude Code (MiniMax M-2.5) & 69 & 0.78 & 74 & 0.59 & 73 & 0.72 & 69 & 0.58 & 69 & 0.16 & 83 & 0.63 \\
\midrule
\multicolumn{13}{l}{\textbf{Kernel-Specialized Agents}} \\
\midrule
AKO4all (Opus-4.6) & 91 & \underline{1.00} & 88 & 1.09 & 88 & \textbf{1.08} & 80 & \underline{1.07} & 84 & 0.30 & 86 & \textbf{1.12} \\
CUDA Opt. Skill (MiniMax M-2.5) & 63 & 0.81 & 64 & 0.77 & 65 & 0.81 & 58 & 0.79 & 53 & 0.21 & 67 & 0.77 \\
\midrule
AutoKernel (GLM-5.0) & 87 & \underline{1.00} & 56 & 1.01 & 64 & 0.99 & 25 & 1.01 & 53 & \textbf{0.82} & 59 & 1.00 \\
AutoKernel (Qwen3.5-27b) & 69 & \underline{1.00} & 75 & 1.03 & 65 & \underline{1.00} & 21 & 1.01 & 40 & 0.37 & 74 & \underline{1.04} \\
AutoKernel (MiniMax M-2.5) & 66 & 0.87 & 71 & \textbf{1.36} & 66 & 0.99 & 50 & 1.02 & 61 & \underline{0.66} & 71 & \underline{1.04} \\
\bottomrule
\end{tabular}%
}
\end{table}

\textbf{Finding 2: Target-language portability does not imply generation-system portability.}
The same Triton task interface masks sharply different system behavior. Claude Code (Opus-4.6) remains relatively accurate across platforms (83--96\%), but Platform A reduces its speedup from 0.86$\times$ on NVIDIA to 0.18$\times$. Iluvatar CoreX exposes the converse failure mode: Claude Code (Qwen3.5-27b) reaches 23\% accuracy, and AutoKernel variants reach only 21--25\%. AutoKernel (GLM-5.0), for example, falls from 87\% on NVIDIA to 25\% on Iluvatar CoreX. Source-level compatibility is therefore insufficient evidence of deployment readiness: the model, scaffold, compiler, runtime, and backend form a coupled generation system whose portability must be measured end to end.

Portability failures are not merely uniform accuracy shifts. Across the 15 configurations shared by all six platforms, Spearman rank correlation with NVIDIA accuracy is 0.87 on Moore Threads, 0.89 on Hygon, 0.66 on Iluvatar CoreX, 0.86 on Platform A, and 0.84 on Platform B. Iluvatar CoreX therefore changes not only absolute success rates but also the relative ordering of generation systems.

Section~\ref{subsec:trajectory} analyzes the trajectories underlying these portability gaps and separates model-code failures from compiler/runtime and backend limitations.

\begin{figure}[!t]
\centering
\includegraphics[width=\textwidth]{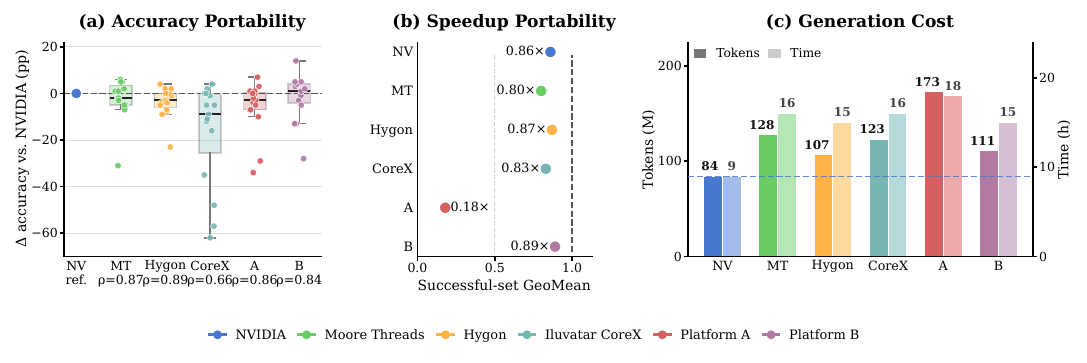}
\caption{Cross-platform portability separates into accuracy, speed, and generation cost. (a) Accuracy changes relative to NVIDIA across 15 shared configurations. NVIDIA is shown once as the zero reference; each non-NVIDIA distribution contains 15 paired configurations; boxes show interquartile ranges, and $\rho$ denotes Spearman rank correlation with NVIDIA. (b,c) Successful-set GeoMean speedup and generation cost for Claude Code (Opus-4.6) on 110 ATen operators; speedup uses each platform's own successful set. Dashed lines mark speedup guides and the NVIDIA cost baseline (84M tokens, 9h).}
\label{fig:cross_platform_portability}
\end{figure}

\textbf{Cross-platform portability separates into accuracy, speed, and generation cost.}
Figure~\ref{fig:cross_platform_portability} complements Table~\ref{tab:cross_platform_main}: it summarizes system-wide accuracy changes and ranking stability, then reports speed and generation cost for a fixed model--scaffold configuration.

\textbf{Finding 3: Hardware heterogeneity imposes an agentic portability tax.}
For a fixed task set, model, and scaffold (110 ATen operators with Claude Code Opus-4.6), we define the descriptive \emph{agentic portability tax} as the additional token or wall-clock cost of a target platform relative to NVIDIA. The five non-NVIDIA platforms require 1.27--2.06$\times$ the tokens and 1.67--2.00$\times$ the time; Platform A reaches 173M tokens and 18 hours, versus 84M and 9 hours on NVIDIA. This tax captures effort omitted by final accuracy and speedup: a system may eventually produce correct kernels while spending substantially more interaction on compilation and runtime failures. It is a property of this controlled configuration, not a universal constant for a vendor or backend.

\subsection{Source-Conditioned Capability Profiles}
\label{subsec:source_profiles}

We treat the 21 configurations in Table~\ref{tab:nvidia_main_ops} as paired observations. Table~\ref{tab:source_profile_summary} summarizes the resulting source-conditioned profiles. Accuracy is defined for all configurations; speedup medians omit the two source--configuration cells with no successful operators.

\begin{table}[!t]
\centering
\caption{Source-conditioned profiles across the 21 configurations in Table~\ref{tab:nvidia_main_ops}. Spd is the median of reported successful-set GeoMeans.}
\label{tab:source_profile_summary}
\small
\begin{tabular}{lccc}
\toprule
\textbf{Source} & \textbf{Median Acc. (\%)} & \textbf{Median Spd ($\times$)} & \textbf{Paired comparison with ATen} \\
\midrule
ATen   & 63 & 0.83 & --- \\
vLLM   & 26 & 1.24 & Acc. $<$ ATen in 20/21 configs. \\
cuBLAS & 36 & 0.53 & Spd $<$ ATen in 18/20 defined pairs \\
\bottomrule
\end{tabular}
\end{table}

Paired comparisons sharpen this result. vLLM is not uniformly ``harder'': its median accuracy is 37~pp below ATen, while its median successful-set speedup is higher, and it exceeds ATen speedup in 15 of 20 defined pairs. cuBLAS presents the reverse boundary: correctness can remain high, but its speedup is below ATen in 18 of 20 defined pairs. Thus, source changes the limiting capability: semantic reconstruction for vLLM versus optimization against a vendor-tuned cuBLAS ceiling.

To test whether these aggregate differences reflect stable operator-level structure, we analyze pre-anti-hack R0 verification outcomes from Claude Code and OpenCode paired with the same two foundation models, Opus-4.6 and GLM-5.0. Across the common 210-operator task universe, 114 operators are solved by all four configurations, whereas 20 are solved by none; the remaining 76 separate method-dependent from broadly stable difficulty. The source split is especially diagnostic: all four solve 75/110 ATen, 18/50 vLLM, and 21/50 cuBLAS operators, while none solve 5/110, 13/50, and 2/50, respectively. The lower vLLM accuracy persists across configurations: it contains the largest stable-hard subset (26\%), including paged attention, quantized GEMM, and MoE primitives. Matched-set comparisons further show that scaffold effects vary by source: Claude Code and OpenCode are statistically indistinguishable with Opus-4.6, whereas with GLM-5.0 Claude Code is faster on shared ATen successes (0.88$\times$ vs. 0.74$\times$) but OpenCode is faster on shared vLLM successes (1.12$\times$ vs. 0.84$\times$). Paired bootstrap confirms this source-dependent reversal for GLM-5.0: the OpenCode/Claude Code speedup ratio is 0.84 (95\% CI [0.71, 0.98]) on ATen but 1.33 (95\% CI [1.07, 1.73]) on vLLM. Appendix~\ref{app:operator_difficulty} reports the complete difficulty and matched-set results, while Appendix~\ref{app:accuracy_speedup_gap} visualizes the related accuracy--speedup compression.

\subsection{Why Portability Fails: Trajectory Analysis}
\label{subsec:trajectory}

Trajectory auditing separates platform-independent model-code failures from platform-induced compiler, API, language-feature, dtype, and runtime incompatibilities. On Moore Threads, mixed-\texttt{int32}/\texttt{int64} type rejection recurs in 288 matched trajectory records. On Iluvatar CoreX, explicit API, language-feature, or dtype incompatibilities affect at least 10 of 110 operators (9.1\%), in addition to compiler hangs and crashes. These results show that the portability gap is partly a platform-interface problem rather than purely a model-capability deficit. Appendix~\ref{app:platform_failures} provides the platform-specific operators, signatures, record counts, and attribution layers.

\subsection{Agentic Scaffolds: Gains, Costs, and Search Behavior}
\label{subsec:cost}

\begin{figure*}[!t]
\centering
\begin{minipage}[c]{0.52\textwidth}
\centering
\captionof{table}{\small Agentic cost measurements on NVIDIA A100.}
\label{tab:nvidia_cost_ops}
\setlength{\tabcolsep}{2.5pt}
\scriptsize
\resizebox{\linewidth}{!}{%
\begin{tabular}{l | c c c}
\toprule
\textbf{Method} & \textbf{Tok. (M)} & \textbf{Tok./Succ. (M)} & \textbf{Time (h)} \\
\midrule
Claude Code (Opus-4.6)      & 263 & 1.45 & 33 \\
Claude Code (GLM-5.0)       & 383 & 2.72 & 48 \\
Claude Code (Qwen3.5-27b)   & 381 & 2.91 & 48 \\
Claude Code (MiniMax M-2.5) & 340 & 3.33 & 50 \\
\midrule
AKO4all (Opus-4.6)              & 904 & 5.20 & 83 \\
CUDA Opt. Skill (MiniMax M-2.5) & 594 & 6.25 & 97 \\
AutoKernel (GLM-5.0)            & 471 & 3.14 & 102 \\
AutoKernel (Qwen3.5-27b)        & 475 & 4.80 & 102 \\
AutoKernel (MiniMax M-2.5)      & 508 & 5.58 & 105 \\
\bottomrule
\end{tabular}}
\end{minipage}\hfill
\begin{minipage}[c]{0.45\textwidth}
\centering
\includegraphics[width=\linewidth]{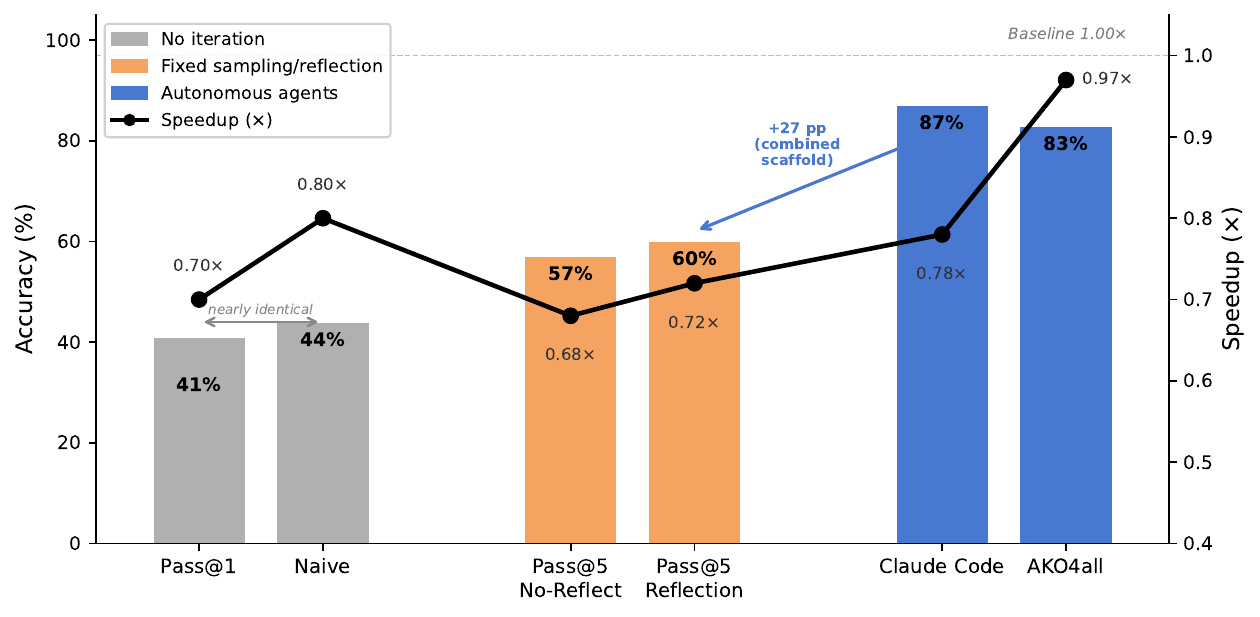}
\captionof{figure}{\small Accuracy and successful-set speedup across generation and debugging regimes (Opus-4.6, NVIDIA A100; 210 operators).}
\label{fig:ablation_overall}
\end{minipage}
\end{figure*}

Table~\ref{tab:nvidia_cost_ops} shows that stronger scaffolds trade capability for cost. AKO4all consumes 904M tokens, 3.44$\times$ Claude Code (Opus-4.6); CUDA Optimized Skill~\cite{cuda_optimized_skill} reaches 6.25M tokens per success; and AutoKernel variants require up to 105 hours. Specialized agents average 4.99M tokens per successful operator, versus 1.45--3.33M for the evaluated Claude Code configurations, partly because they continue searching after correctness to optimize performance. Accuracy and speedup alone can therefore reward an operationally impractical search policy; tokens per success and wall time show the cost of these gains. 

\label{subsec:ablation}

We compare increasingly interactive generation regimes in Figure~\ref{fig:ablation_overall}, with per-source results reported in Appendix~\ref{app:ablation_full}. Autonomous interactive debugging produces the largest observed accuracy gain: Claude Code reaches 87\%, 27~pp above fixed traceback-driven reflection, although the comparison changes multiple scaffold components simultaneously. Fixed reflection also exhibits an \emph{exploitation--exploration trap}: it underperforms independent sampling on vLLM (24\% vs.\ 28\%), while among 21 jointly failed operators it produces substantially higher cross-turn code similarity (88\% vs.\ 36\%), consistent with local patching rather than broader search (Appendix~\ref{app:reflection_similarity}).

\setcounter{topnumber}{2}

\section{Conclusion}
\label{sec:conclusion}

We introduced \textbf{KernelGenBench}, the first unified multi-source $\times$ multi-chip benchmark infrastructure. Its common Triton target spans six hardware platforms, providing the broadest cross-vendor hardware coverage among existing kernel-generation benchmarks. Its controlled MS and MC views cover 210 operators from three sources on NVIDIA and a semantically stable 110-operator set across six platforms, respectively. The results establish operator source, hardware platform, and agentic scaffold as distinct capability dimensions: source changes the correctness and optimization challenge, Triton tasks can transfer poorly across platforms, and more interactive scaffolds achieve higher correctness at high cost.

\textbf{Limitations and Future Work.} First, the infrastructure implements cross-vendor source--platform paths beyond the reported MS and MC views, but production-runtime and vendor-library references are not uniformly public or semantically matched; future work will broaden evaluation as aligned references become available. Second, costly agent configurations mostly use a single trajectory because exhaustive repetition is expensive. We will curate a high-quality representative subset for repeated evaluation under practical budgets. Third, two platforms remain anonymous under commercial agreements. Future work will add new vendors to KernelGenBench and provide access to the anonymous platforms through jointly released leaderboards while preserving confidentiality.

\clearpage
\begin{ack}
The authors would like to thank BAAI for providing the computational resources and API tokens that supported this research. We are also grateful to Yang Yu and Liangdong Wang for their insightful discussions and valuable assistance throughout the development of this work. Special thanks go to Yipeng Jiang for her contribution to the illustrations and the overview figure in this paper. Additionally, we express our sincere gratitude to the participating hardware vendors for their continuous communication, technical feedback, and collaborative support in co-building this benchmarking ecosystem.
\end{ack}

\bibliographystyle{unsrtnat} 
\bibliography{main} 

\newpage
\appendix
\appendix

\section{Prompt Design}
\label{app:prompt_construction}

KernelGenBench dynamically constructs structured prompts for each operator rather than relying on vague natural-language descriptions. Table~\ref{tab:prompt_modules} summarizes the prompt template modules and their differences between the LLM Sampling and Agentic evaluation paradigms.

\begin{table}[htbp]
\centering
\caption{Prompt template modules: LLM Sampling vs.\ Agentic Frameworks.}
\label{tab:prompt_modules}
\small
\begin{tabular}{lcc}
\toprule
\textbf{Module} & \textbf{LLM Sampling} & \textbf{Agentic} \\
\midrule
System Role & \checkmark & \checkmark \\
One-shot Example & \checkmark & \checkmark \\
Operator Spec / Baseline Source Injection & \checkmark & \checkmark \\
Function Signature \& Type Constraints & \checkmark & \checkmark \\
Anti-Cheat Rules & \checkmark & \checkmark \\
Testing Environment Description & \checkmark & \checkmark \\
Device-Specific Constraints (Heterogeneous) & \checkmark & \checkmark \\
\midrule
Accelerator Device Assignment (Device ID) & --- & \checkmark \\
Verification CLI Command & --- & \checkmark \\
Iterative Workflow Guide & --- & \checkmark \\
Performance Optimization Guide & --- & \checkmark \\
\bottomrule
\end{tabular}
\end{table}

The key difference across the three subsets lies in the ``Operator Spec / Baseline Source Injection'' module. In no subset is the underlying production kernel implementation exposed to the model. Table~\ref{tab:source_context} distinguishes the interface-level context that is provided from the implementation that remains hidden.

\begin{table}[htbp]
\centering
\caption{Source-specific context exposed during prompt construction. Python wrapper or harness source describes how an interface is invoked; it is not the underlying production kernel implementation.}
\label{tab:source_context}
\small
\begin{tabular}{p{0.11\textwidth} p{0.39\textwidth} p{0.37\textwidth}}
\toprule
\textbf{Subset} & \textbf{Provided to the model} & \textbf{Not provided} \\
\midrule
ATen & Runtime \texttt{FunctionSchema}, overload variants, wrapper signatures, argument and return specifications, and available docstrings. & PyTorch's native C++/CUDA operator implementation. \\
\addlinespace
vLLM & Source of the benchmark's Python baseline wrapper, including the exact signature, argument order, in-place behavior, and invocation of the corresponding \texttt{\_custom\_ops} interface. & The CUDA/C++ implementation behind the vLLM custom operator. \\
\addlinespace
cuBLAS & Source of the benchmark's Python/\texttt{ctypes} baseline wrapper and interface metadata, including the C-API signature and call sequence used by the harness. & NVIDIA's proprietary cuBLAS implementation. \\
\bottomrule
\end{tabular}
\end{table}

The prompts were designed by HPC engineers to reflect a realistic kernel-development workflow. Within each evaluation setting, every operator follows the same structured construction protocol, with schemas, signatures, wrapper context, verification commands, and device constraints populated automatically. We do not manually tune a prompt for an individual operator or retrospectively select the best-performing formulation. The complete templates and construction code are available in the public repository.

Figure~\ref{fig:prompt_comparison} uses \texttt{aten::cos} as an example, showing side-by-side the actual prompts under the LLM Sampling and Agentic paradigms (shared portions abbreviated as \texttt{[...]} for brevity). The prompts shown are for NVIDIA; on alternative platforms, the device-specific references (e.g., ``GPU programmer'') are automatically replaced with the corresponding platform-specific terminology, and additional device constraints are appended.

\begin{figure*}[htbp]
\centering
\begin{minipage}[t]{0.48\textwidth}
\centering
\textbf{LLM Sampling Prompt}
\begin{tcolorbox}[colback=gray!5, colframe=gray!60, fontupper=\ttfamily\scriptsize, left=2pt, right=2pt, top=2pt, bottom=2pt, boxrule=0.5pt]
You are a skilled GPU programmer proficient in Triton. Your task is to generate a Triton kernel function.\\[3pt]
Here is an example of a PyTorch function and its corresponding Triton kernel implementation:\\
\textnormal{\textit{[one-shot add example, with kernel + wrapper]}}\\[3pt]
You must strictly adhere to the following specifications:\\
The Triton kernel should implement:\\
\texttt{cos = torch.ops.aten.cos}\\[3pt]
The ATen operators involved are:\\
~~- cos\\
~~- cos.out\\
~~- cos.int\\
~~- \textnormal{\textit{[... 6 overload variants in total]}}\\[3pt]
The Python wrapper functions should have signatures like:\\
~~- def cos(...)\\
~~- def cos\_out(...)\\
~~- \textnormal{\textit{[...]}}\\[3pt]
Input and Output Args:\\
~~aten::cos(Tensor self) -> Tensor\\
~~\textnormal{\textit{[... full signature for each variant]}}\\[3pt]
CRITICAL REQUIREMENTS:\\
1. Handle all overload variants correctly\\
2. Handle non-contiguous tensors correctly\\[3pt]
Generate valid Triton code directly without any explanations. Use \texttt{```python```} format.
\end{tcolorbox}
\end{minipage}%
\hfill
\begin{minipage}[t]{0.48\textwidth}
\centering
\textbf{Agentic Framework Prompt}
\begin{tcolorbox}[colback=blue!3, colframe=blue!40, fontupper=\ttfamily\scriptsize, left=2pt, right=2pt, top=2pt, bottom=2pt, boxrule=0.5pt]
\# Triton Kernel Implementation Task\\
Operator: cos ~~ Full name: aten::cos\\
GPU ID: 0\\[3pt]
\#\# Runtime Environment\\
CUDA\_VISIBLE\_DEVICES=0\\[3pt]
\#\# Operator Specification\\
\textnormal{\textit{[same schema and signature as LLM prompt]}}\\[3pt]
\#\# Example\\
\textnormal{\textit{[same one-shot add example as LLM prompt]}}\\[3pt]
\#\# Verification Tool\\
\texttt{python verify\_single.py --code kernel.py}\\
\texttt{~~--op aten::cos --output-json}\\[3pt]
\#\# Iterative Workflow\\
1. Write implementation to \texttt{kernel.py}\\
2. Run verification command\\
3. If failed, read error and fix\\
4. Repeat until all tests pass\\[3pt]
\#\# Performance Optimization\\
After correctness is achieved:\\
- Tune BLOCK\_SIZE, num\_warps, num\_stages\\
- Re-run verification to confirm speedup\\[3pt]
\#\# Anti-Cheat Rules\\
\textnormal{\textit{[same rules as LLM prompt]}}
\end{tcolorbox}
\end{minipage}
\caption{Side-by-side comparison of LLM Sampling and Agentic prompts for \texttt{aten::cos}. Shared content is abbreviated with \texttt{[...]}.}
\label{fig:prompt_comparison}
\end{figure*}

\section{Full Operator List}
\label{app:full_op_list}

Tables~\ref{tab:aten_ops}, \ref{tab:vllm_ops}, and \ref{tab:cublas_ops} list the complete operator names for the three KernelGenBench subsets.

\begin{table}[htbp]
\centering
\caption{Full operator list for the ATen subset (110 operators).}
\label{tab:aten_ops}
\footnotesize
\setlength{\tabcolsep}{3pt}
\resizebox{\textwidth}{!}{%
\begin{tabular}{llll}
\toprule
\texttt{\_index\_put\_impl\_} & \texttt{\_local\_scalar\_dense} & \texttt{\_softmax}              & \texttt{binary\_cross\_entropy\_with\_logits} \\
\texttt{\_to\_copy}          & \texttt{acosh}                  & \texttt{add}                    & \texttt{affine\_grid\_generator} \\
\texttt{add\_}               & \texttt{amin}                   & \texttt{arange}                 & \texttt{upsample\_nearest2d\_backward} \\
\texttt{argmax}              & \texttt{as\_strided}            & \texttt{asin}                   & \texttt{log\_sigmoid\_backward} \\
\texttt{bernoulli}           & \texttt{bitwise\_not}           & \texttt{bmm}                    & \texttt{reflection\_pad1d\_backward} \\
\texttt{cat}                 & \texttt{clone}                  & \texttt{contiguous}             & \texttt{rrelu\_with\_noise\_backward} \\
\texttt{copy\_}              & \texttt{cos}                    & \texttt{cosh}                   & \texttt{unsafe\_split\_with\_sizes} \\
\texttt{cumsum}              & \texttt{diff}                   & \texttt{div}                    & \texttt{smooth\_l1\_loss\_backward} \\
\texttt{div\_}               & \texttt{embedding}              & \texttt{empty\_strided}         & \texttt{margin\_ranking\_loss} \\
\texttt{eq}                  & \texttt{erfc}                   & \texttt{expand}                 & \texttt{rrelu\_with\_noise} \\
\texttt{expand\_as}          & \texttt{exponential\_}          & \texttt{fill\_}                 & \texttt{pairwise\_distance} \\
\texttt{floor}               & \texttt{floor\_divide}          & \texttt{fmax}                   & \texttt{softplus\_backward} \\
\texttt{full}                & \texttt{gather}                 & \texttt{gt}                     & \texttt{new\_empty\_strided} \\
\texttt{hardsigmoid}         & \texttt{heaviside}              & \texttt{huber\_loss}            & \texttt{soft\_margin\_loss} \\
\texttt{i0}                  & \texttt{im2col}                 & \texttt{index}                  & \texttt{select\_backward} \\
\texttt{index\_put\_}        & \texttt{index\_select}          & \texttt{item}                   & \texttt{mish\_backward} \\
\texttt{le}                  & \texttt{linear}                 & \texttt{log10}                  & \texttt{masked\_fill\_} \\
\texttt{logaddexp2}          & \texttt{logit}                  & \texttt{matmul}                 & \texttt{resolve\_conj} \\
\texttt{mean}                & \texttt{mish}                   & \texttt{mm}                     & \texttt{resolve\_neg} \\
\texttt{mul}                 & \texttt{narrow}                 & \texttt{neg}                    & \texttt{special\_entr} \\
\texttt{new\_ones}           & \texttt{ones\_like}             & \texttt{poisson}                & \texttt{scalar\_tensor} \\
\texttt{polygamma}           & \texttt{pow}                    & \texttt{prelu}                  & \texttt{unsafe\_split} \\
\texttt{renorm}              & \texttt{reshape}                & \texttt{rot90}                  & \texttt{unsqueeze} \\
\texttt{rsqrt}               & \texttt{rsub}                   & \texttt{scatter}                & \texttt{zeros\_like} \\
\texttt{select}              & \texttt{sgn}                    & \texttt{silu}                   & \texttt{softmax} \\
\texttt{sin}                 & \texttt{sort}                   & \texttt{square}                 & \texttt{stack} \\
\texttt{sub}                 & \texttt{sum}                    & \texttt{t}                      & \texttt{to} \\
\texttt{zero\_}              & \texttt{zeros}                  &                                 & \\
\bottomrule
\end{tabular}
}
\end{table}

\begin{table}[htbp]
\centering
\caption{Full operator list for the vLLM subset (50 operators, from vLLM v0.13.0).}
\label{tab:vllm_ops}
\small
\resizebox{\textwidth}{!}{%
\begin{tabular}{lll}
\toprule
\texttt{allspark\_w8a16\_gemm} & \texttt{apply\_repetition\_penalties\_cuda} & \texttt{awq\_gemm} \\
\texttt{awq\_marlin\_moe\_repack} & \texttt{batched\_moe\_align\_block\_size} & \texttt{concat\_and\_cache\_mla} \\
\texttt{convert\_fp8} & \texttt{convert\_vertical\_slash\_indexes} & \texttt{copy\_blocks} \\
\texttt{copy\_blocks\_mla} & \texttt{cp\_gather\_cache} & \texttt{cp\_gather\_indexer\_k\_quant\_cache} \\
\texttt{cutlass\_pack\_scale\_fp8} & \texttt{cutlass\_scaled\_mm} & \texttt{cutlass\_scaled\_mm\_azp} \\
\texttt{fused\_add\_rms\_norm} & \texttt{fused\_qk\_norm\_rope} & \texttt{gather\_and\_maybe\_dequant\_cache} \\
\texttt{ggml\_dequantize} & \texttt{ggml\_moe\_a8} & \texttt{ggml\_moe\_a8\_vec} \\
\texttt{ggml\_mul\_mat\_a8} & \texttt{ggml\_mul\_mat\_vec\_a8} & \texttt{gptq\_gemm} \\
\texttt{gptq\_marlin\_24\_gemm} & \texttt{gptq\_marlin\_gemm} & \texttt{gptq\_marlin\_moe\_repack} \\
\texttt{gptq\_shuffle} & \texttt{grouped\_topk} & \texttt{hadacore\_transform} \\
\texttt{marlin\_int4\_fp8\_preprocess} & \texttt{merge\_attn\_states} & \texttt{moe\_align\_block\_size} \\
\texttt{moe\_lora\_align\_block\_size} & \texttt{moe\_sum} & \texttt{paged\_attention\_v1} \\
\texttt{paged\_attention\_v2} & \texttt{permute\_cols} & \texttt{reshape\_and\_cache} \\
\texttt{reshape\_and\_cache\_flash} & \texttt{rms\_norm} & \texttt{rms\_norm\_dynamic\_per\_token\_quant} \\
\texttt{rms\_norm\_per\_block\_quant} & \texttt{rotary\_embedding} & \texttt{scaled\_fp8\_quant} \\
\texttt{scaled\_int8\_quant} & \texttt{selective\_scan\_fwd} & \texttt{shuffle\_rows} \\
\texttt{swap\_blocks} & \texttt{topk\_softmax} & \\
\bottomrule
\end{tabular}
}
\end{table}

\begin{table}[htbp]
\centering
\caption{Full operator list for the cuBLAS subset (50 operators, based on cuBLAS 12.4).}
\label{tab:cublas_ops}
\small
\resizebox{\textwidth}{!}{%
\begin{tabular}{lll}
\toprule
\texttt{cublasCcopy\_v2} & \texttt{cublasCdotu\_v2} & \texttt{cublasCgemmStridedBatched} \\
\texttt{cublasCgemmStridedBatched\_64} & \texttt{cublasCgemm\_v2} & \texttt{cublasCgemvBatched\_64} \\
\texttt{cublasCgemvStridedBatched} & \texttt{cublasCgemv\_v2} & \texttt{cublasCgeru\_v2} \\
\texttt{cublasCsymm\_v2} & \texttt{cublasCsymv\_v2} & \texttt{cublasCsyrkEx} \\
\texttt{cublasDasum\_v2} & \texttt{cublasDaxpy\_v2} & \texttt{cublasDcopy\_v2} \\
\texttt{cublasDgemmBatched} & \texttt{cublasDgemmStridedBatched} & \texttt{cublasDgemmStridedBatched\_64} \\
\texttt{cublasDgemvBatched} & \texttt{cublasDgemvStridedBatched} & \texttt{cublasDgemv\_v2} \\
\texttt{cublasDsbmv\_v2} & \texttt{cublasDsyr2\_v2} & \texttt{cublasDtrsmBatched} \\
\texttt{cublasHgemmBatched} & \texttt{cublasHgemmStridedBatched} & \texttt{cublasSaxpy\_v2} \\
\texttt{cublasSdgmm} & \texttt{cublasSdot\_v2} & \texttt{cublasSgeam} \\
\texttt{cublasSgemmBatched\_64} & \texttt{cublasSgemmEx} & \texttt{cublasSgemmStridedBatched} \\
\texttt{cublasSgemm\_v2} & \texttt{cublasSgemvBatched} & \texttt{cublasSgemvStridedBatched} \\
\texttt{cublasSger\_v2} & \texttt{cublasSscal\_v2} & \texttt{cublasSsyrk\_v2} \\
\texttt{cublasStbmv\_v2} & \texttt{cublasStrsm\_v2} & \texttt{cublasStrsv\_v2} \\
\texttt{cublasZdotc\_v2} & \texttt{cublasZgemmBatched} & \texttt{cublasZgemmStridedBatched} \\
\texttt{cublasZgemvBatched} & \texttt{cublasZgemvStridedBatched} & \texttt{cublasZgerc\_v2} \\
\texttt{cublasZswap\_v2} & \texttt{cublasZtrsmBatched} & \\
\bottomrule
\end{tabular}
}
\end{table}

\FloatBarrier
\section{cuBLAS GEMM Precision Variant Coverage}
\label{app:cublas_gemm}

Table~\ref{tab:cublas_gemm_app} shows the 14 independent problems derived from GEMM (General Matrix Multiplication) operators in the cuBLAS subset, broken down by data precision, batching mode, and index width. This illustrates the fine-grained API fragmentation typical of closed-source low-level libraries.

\begin{table}[htbp]
\centering
\caption{Fine-grained cuBLAS operator variants: GEMM (General Matrix Multiplication) as an example.}
\label{tab:cublas_gemm_app}
\resizebox{\textwidth}{!}{
\begin{tabular}{lccccc}
\toprule
\textbf{Precision} & \textbf{Standard (\_v2)} & \textbf{StridedBatched} & \textbf{Batched} & \textbf{64-bit Index (\_64)} & \textbf{Other Variants} \\
\midrule
\textbf{Float32 (S)} & cublasSgemm\_v2 & cublasSgemmStridedBatched & --- & cublasSgemmBatched\_64 & cublasSgemmEx \\
\textbf{Float64 (D)} & --- & cublasDgemmStridedBatched & cublasDgemmBatched & cublasDgemmStridedBatched\_64 & --- \\
\textbf{Complex64 (C)} & cublasCgemm\_v2 & cublasCgemmStridedBatched & --- & cublasCgemmStridedBatched\_64 & --- \\
\textbf{Complex128 (Z)} & --- & cublasZgemmStridedBatched & cublasZgemmBatched & --- & --- \\
\textbf{Float16 (H)} & --- & cublasHgemmStridedBatched & cublasHgemmBatched & --- & --- \\
\bottomrule
\end{tabular}
}
\end{table}

\section{Cross-Platform Prompt and Tolerance Configuration}
\label{app:sandbox_chip}

This section details the hardware-specific configurations injected by KernelGenBench-MC for heterogeneous hardware-platform evaluation.

\textbf{Hardware Configuration.}
Table~\ref{tab:hardware_software_config} summarizes the accelerator used on each platform and its evaluation environment. Platforms A and B remain de-identified under their confidentiality requirements.

\begin{table*}[htbp]
\centering
\caption{Hardware and software configurations for KernelGenBench-MC. All evaluations use a full node and the official image provided by the corresponding vendor.}
\label{tab:hardware_software_config}
\small
\begin{tabularx}{\textwidth}{@{}p{0.17\textwidth} p{0.20\textwidth} p{0.19\textwidth} X@{}}
\toprule
\textbf{Paper label} & \textbf{Accelerator} & \textbf{Execution stack} & \textbf{Key software versions} \\
\midrule
NVIDIA & NVIDIA A100 & CUDA & cuBLAS 12.4; vLLM 0.13.0 \\
Moore Threads & MTT S5000 & MUSA / MTGPU & PyTorch 2.5.0; torch\_musa 2.5.0; MUSA 3.1; Triton 3.1.0 (vendor backend) \\
Hygon & DCU BW series (gfx936) & DTK / HIP--ROCm & DTK 25.04; Triton 3.0.0+das.opt4.dtk2504 \\
Iluvatar CoreX & BI-V150 & IXUCA & Ubuntu 22.04.5; PyTorch 2.7.1; IXUCA 4.4.0; Triton 3.1.0 (vendor build) \\
Platform A & Not disclosed & Not disclosed & Not disclosed under commercial confidentiality \\
Platform B & Not disclosed & Not disclosed & Not disclosed under commercial confidentiality \\
\bottomrule
\end{tabularx}
\end{table*}

\textbf{Cross-Vendor Reference Adapters.} KernelGenBench provides completed source--platform adapter paths beyond the two analytical views reported in the main paper. Table~\ref{tab:cross_vendor_adapters} records the public production-runtime and vendor-library mappings used by this infrastructure. All public mappings were validated on their target platforms. These paths also support ongoing vendor-collaborative kernel-development leaderboards, providing a common submission and verification interface for contributing high-quality operators; Platform A and Platform B mappings remain de-identified.

\begin{table*}[htbp]
\centering
\caption{Cross-vendor production-runtime and library-reference mappings in KernelGenBench. The completed adapter layer supports source--platform evaluation and vendor-collaborative kernel development; Platform A and Platform B retain de-identified mappings.}
\label{tab:cross_vendor_adapters}
\small
\begin{tabularx}{\textwidth}{@{}p{0.16\textwidth} p{0.20\textwidth} p{0.23\textwidth} X@{}}
\toprule
\textbf{Platform} & \textbf{Execution stack} & \textbf{Production-runtime path} & \textbf{Vendor-library path} \\
\midrule
NVIDIA & CUDA & Upstream vLLM & cuBLAS (native interface) \\ \addlinespace[3pt]
\mbox{Moore Threads} & \mbox{MUSA / MTGPU} & vLLM-MUSA & muBLAS (cuBLAS-style mapping) \\ \addlinespace[3pt]
Hygon & \mbox{DTK / HIP--ROCm} & vLLM on DTK & hipBLAS (cuBLAS-style mapping) \\ \addlinespace[3pt]
\mbox{Iluvatar CoreX} & IXUCA & CoreX vLLM distribution & CoreX cuBLAS-compatible library \\ \addlinespace[3pt]
Platform A & De-identified & De-identified vLLM/runtime adapter & De-identified vendor-library adapter \\ \addlinespace[3pt]
Platform B & De-identified & De-identified vLLM/runtime adapter & De-identified vendor-library adapter \\
\bottomrule
\end{tabularx}
\end{table*}

\textbf{Hardware-Aware Prompt Templates.}
The framework maintains separate prompt template directories per device type (e.g., \texttt{templates/platform\_a/}, \texttt{templates/platform\_b/}), automatically selecting and assembling the corresponding template at runtime based on detected hardware (e.g., injecting platform-specific import statements and runtime initialization code).

\FloatBarrier
\section{Distributed Sandbox Infrastructure}
\label{app:sandbox_infra}

This section details the production-grade dispatch registration and the process isolation and distributed scheduling infrastructure underlying KernelGenBench's full-node parallel evaluation.

\textbf{Production-Grade Dispatch Registration.}
For ATen operators, the system uses \texttt{register\_scanner.py} to perform AST scanning and capture \texttt{@register} decorators, force-mounting Triton kernels into PyTorch's \texttt{torch.library} dispatch tree. All ATen operators are therefore tested under PyTorch's full low-level dispatch mechanism, not isolated Python function calls.

\textbf{Full-Node Parallel Evaluation with Process Isolation.}
To enable full-node utilization, KernelGenBench dispatches all operators in parallel across all available accelerator devices on the machine. Each operator runs in a fully isolated subprocess (\texttt{multiprocessing spawn}), with its own workspace directory and exclusive device assignment, ensuring that a crash, timeout, or device memory corruption in one operator cannot affect any other. The distributed resource scheduler \texttt{DeviceManager} acquires hardware control via kernel-level file locks, recording the holder's PID and automatically reclaiming locks from crashed processes to prevent permanent device resource leaks. When a subprocess triggers a timeout (300s for correctness, 600s for performance), the parent process sends \texttt{SIGKILL} to destroy the entire process tree.

\textbf{Adaptive Numerical Tolerance.}
Numerical comparison of low-level operators is sensitive to floating-point truncation errors. The comparison uses $\text{atol}=10^{-4}D_{\text{reduce}}$ and a dtype-specific relative tolerance. By default, $D_{\text{reduce}}=1$, providing a fixed absolute tolerance of $10^{-4}$; this default is used for non-accumulating comparisons and normalized reductions such as \texttt{softmax} and \texttt{mean}. Tests whose outputs directly accumulate multiple terms, such as sums and matrix products, explicitly pass the corresponding accumulation length as $D_{\text{reduce}}$. The relative tolerance follows dtype effective-bit resolution ($\text{FP32} \to 10^{-5}, \text{FP16} \to 10^{-3}, \text{BF16} \to 0.016$). For MC, L1 static scanning and L2 dual-execution comparison form the common enforcement protocol on every platform. NVIDIA additionally uses the L3 GPU-profiler fingerprint; equivalent profiler interfaces remain unavailable across the other hardware stacks. L3 is therefore a supplementary NVIDIA audit rather than a prerequisite for the cross-platform correctness decision. The representative shortcut behaviors in Table~\ref{tab:antihack_cases} are rejected by the shared L1/L2 checks.

\textbf{Platform-local reference policy.} KernelGenBench-MC evaluates the fixed 110-task ATen suite on every platform. Each task is checked against the native ATen reference available in that platform's local PyTorch backend. If this reference API is unavailable, the task remains in the 110-task denominator and is recorded as unsuccessful. MC accuracy therefore measures end-to-end platform success across generation, compilation, local-reference verification, and execution, rather than generated-kernel correctness in isolation.

\section{Evaluation Pipeline Algorithm Details}
\label{app:pipeline_algorithms}

This section provides the full pseudocode for both the agent dispatch pipeline and the LLM Sampling Methods generation pipeline described in Section~\ref{subsec:pipeline_and_antihack}.

\textbf{Common External Budget and Method-Specific Resources.}
All evaluated methods receive the same target-operator specification and test suite, and every final submission is judged by the same correctness and applicable anti-hack checks under a common 30-minute wall-clock budget per operator task. We control this task-level budget rather than the number of internal iterations: agent frameworks define an iteration differently and autonomously allocate their time among generation, debugging, verification, and optimization. Pass@k and vanilla agentic frameworks do not receive profiling access from the benchmark. Some kernel-specialized agents include profiling, skills, or hardware tools as intrinsic parts of their published workflows; these capabilities are retained because the evaluated object is the complete kernel-generation system. Thus, the comparison standardizes external task conditions and final evaluation, while intentionally preserving method-specific mechanisms. We report token use and wall-clock cost to capture the computational consequences of these method-specific search strategies.

\begin{algorithm}[htbp]
\caption{Vanilla Agentic Framework Evaluation Dispatch}
\label{alg:agent_pipeline}
\begin{algorithmic}[1]
\REQUIRE Operator set $\mathcal{O}$, agent backend $M$, device pool $\mathcal{G}$, timeout $T_{\max}$
\STATE $\mathcal{C} \leftarrow \emptyset$ \COMMENT{Final candidate kernel for each operator}
\FOR{\textbf{each} $op \in \mathcal{O}$ \textbf{in parallel} (bounded by $|\mathcal{G}|$)}
    \STATE $dev \leftarrow \text{DeviceManager.acquire}(\mathcal{G})$ \COMMENT{Works identically on all platforms}
    \STATE $ws \leftarrow \text{CreateWorkspace}(op)$
    \STATE $prompt \leftarrow \text{BuildPrompt}(op, dev, \texttt{verify\_cmd})$
    \STATE $proc \leftarrow M.\text{launch}(prompt, ws, dev)$ \COMMENT{Spawn agent process}
    \STATE \COMMENT{Agent autonomously writes, verifies, debugs, and optimizes the kernel}
    \STATE \textbf{wait} $proc$ until completion or $T_{\max}$
    \STATE $code \leftarrow M.\text{finish}(proc, ws)$ \COMMENT{Extract final kernel}
    \STATE $\mathcal{C}[op] \leftarrow code$
    \STATE $\text{DeviceManager.release}(dev)$
\ENDFOR
\STATE \textbf{return} $\text{BatchAntiHackCheck}(\mathcal{C})$
\end{algorithmic}
\end{algorithm}

\textbf{Formal definition of the LLM Sampling Methods pipeline:}
For a given operator, the system draws up to $k$ independent samples. Every round reconstructs the prompt from the original operator specification, without including previous candidates, execution errors, or tracebacks. Once a candidate passes correctness verification, the system stores that kernel, locks the operator, and removes it from the remaining task pool. After sampling finishes for all operators, the evaluator applies anti-hack checks once to the complete set of stored kernels. A flagged kernel is counted as unsuccessful without replacement sampling. The overall clean pass rate is the fraction of operators whose stored kernel passes this final filter.

\begin{algorithm}[htbp]
\caption{LLM Sampling Methods: Pass@K Sequential Early-Stopping Evaluation Pipeline}
\label{alg:pass_k}
\begin{algorithmic}[1]
\REQUIRE Operator set $\mathcal{O}$, max sampling rounds $K$, model policy $\pi$, evaluator $\mathcal{E}$
\STATE $\mathcal{O}_{remaining} \leftarrow \mathcal{O}$, $\mathcal{C}_{passed} \leftarrow \emptyset$ \COMMENT{Operator-to-candidate map}
\FOR{$round = 1$ \textbf{to} $K$}
    \FOR{\textbf{each} $op \in \mathcal{O}_{remaining}$}
        \STATE $p \leftarrow \text{ConstructPrompt}(op)$ \COMMENT{Fresh prompt; no feedback from earlier rounds}
        \STATE $c \leftarrow \pi(p)$ \COMMENT{Generate code}
        \STATE $is\_correct, speedup \leftarrow \mathcal{E}.\text{verify}(c)$
        \IF{$is\_correct$}
            \STATE $\mathcal{C}_{passed}[op] \leftarrow (c, speedup)$
        \ENDIF
    \ENDFOR
    \STATE $\mathcal{O}_{remaining} \leftarrow \mathcal{O}_{remaining} \setminus \operatorname{keys}(\mathcal{C}_{passed})$ \COMMENT{Correctness-based early stopping}
    \IF{$\mathcal{O}_{remaining} == \emptyset$}
        \STATE \textbf{break}
    \ENDIF
\ENDFOR
\STATE \textbf{return} $\text{BatchAntiHackCheck}(\mathcal{C}_{passed})$ \COMMENT{Flagged kernels fail; no resampling}
\end{algorithmic}
\end{algorithm}

\section{Stress Test Shape and Dispatch Configurations}
\label{app:sandbox_stress}

This section details the combinatorial test case construction and shape/stride configurations used in KernelGenBench's parametrized test cases. All constants are defined in \texttt{accuracy\_utils.py}.

\textbf{Extreme Tensor Distributions and Boundary Stress Testing.}
Test cases deliberately inject many non-aligned long-tail sizes (e.g., $40999$, odd boundaries constructed as $2^d + 17$), mixed with full-dimensional coverage from scalar \texttt{()} to 5D tensors \texttt{(16, 7, 57, 32, 29)}. Minimal-size tests (e.g., \texttt{shape=(1, 2)}) expose kernel launch overhead degradation.

\textbf{Example: \texttt{softmax} Test Case Construction.}
For \texttt{softmax}, the semantic parameter is \texttt{dim} (reduction axis). The $k_i = 18$ test cases are the Cartesian product of three axes: shape $\in$ \{(1,\,256),\,(4096,\,256),\,(200,\,2560,\,3)\}, dtype $\in$ \{float16,\,float32,\,bfloat16\}, and dim $\in$ \{0,\,1\}. A kernel is counted as passing only if all 18 cases satisfy the adaptive numerical tolerance. The shape and stride grids used across all operator categories are listed in Tables~\ref{tab:test_shapes} and~\ref{tab:test_strides}.

\begin{table}[htbp]
\centering
\caption{Shape configurations by operator category.}
\label{tab:test_shapes}
\small
\resizebox{\columnwidth}{!}{%
\begin{tabular}{ll}
\toprule
\textbf{Category} & \textbf{Shapes} \\
\midrule
Pointwise & \texttt{()}, \texttt{(1,)}, \texttt{(1024, 1024)}, \texttt{(20, 320, 15)}, \texttt{(16, 128, 64, 60)}, \texttt{(16, 7, 57, 32, 29)} \\
Reduction & \texttt{(1, 2)}, \texttt{(4096, 256)}, \texttt{(200, 40999, 3)} \\
Reduction (small) & \texttt{(1, 2)}, \texttt{(4096, 256)}, \texttt{(200, 2560, 3)} \\
BLAS (M, N) & \texttt{(1, 32)}, \texttt{(160, 1024)}, \texttt{(5333, 497)} \\
BLAS (M, N, K) & \texttt{(1, 1, 32)}, \texttt{(15, 160, 1024)}, \texttt{(495, 5333, 71)} \\
1D (power-of-2) & \texttt{(16,)}, \texttt{(64,)}, \texttt{(256,)}, \texttt{(1024,)} \\
1D (non-aligned) & \texttt{(33,)}, \texttt{(81,)}, \texttt{(273,)}, \texttt{(1041,)} \\
Upsample & \texttt{(32,16,128,128)}, \texttt{(15,37,256,256)}, \texttt{(3,5,127,127)}, \texttt{(128,192,42,51)}, \texttt{(3,7,1023,1025)} \\
\bottomrule
\end{tabular}
}
\end{table}

\begin{table}[htbp]
\centering
\caption{Stride configurations for memory layout stress testing.}
\label{tab:test_strides}
\small
\begin{tabular}{lll}
\toprule
\textbf{Layout Type} & \textbf{Shape} & \textbf{Stride} \\
\midrule
\multicolumn{3}{l}{\textit{1D Contiguous}} \\
& \texttt{(1,)} & \texttt{(1,)} \\
& \texttt{(1024,)} & \texttt{(1,)} \\
& \texttt{(1000000,)} & \texttt{(1,)} \\
\midrule
\multicolumn{3}{l}{\textit{1D Dilated}} \\
& \texttt{(1,)} & \texttt{(2,)} \\
& \texttt{(1024,)} & \texttt{(2,)} \\
& \texttt{(1000000,)} & \texttt{(2,)} \\
\midrule
\multicolumn{3}{l}{\textit{2D Contiguous}} \\
& \texttt{(1, 1024)} & \texttt{(1024, 1)} \\
& \texttt{(10000, 128)} & \texttt{(128, 1)} \\
\midrule
\multicolumn{3}{l}{\textit{2D Transposed}} \\
& \texttt{(1024, 1)} & \texttt{(1, 1024)} \\
& \texttt{(128, 10000)} & \texttt{(1, 128)} \\
\midrule
\multicolumn{3}{l}{\textit{3D Contiguous}} \\
& \texttt{(20, 320, 15)} & \texttt{(4800, 15, 1)} \\
& \texttt{(200, 40999, 3)} & \texttt{(122997, 3, 1)} \\
\midrule
\multicolumn{3}{l}{\textit{3D Transposed}} \\
& \texttt{(320, 20, 15)} & \texttt{(15, 4800, 1)} \\
& \texttt{(3, 40999, 200)} & \texttt{(1, 3, 122997)} \\
\midrule
\multicolumn{3}{l}{\textit{5D Irregular}} \\
& \texttt{(10, 10, 10, 10, 10)} & \texttt{(1, 10000, 23, 399, 1024)} \\
\bottomrule
\end{tabular}
\end{table}

\section{Additional Analysis on Generation Behaviors}
\label{app:generation_behaviors}

This section provides supplementary observations regarding model-specific generation bottlenecks and anti-hack interception statistics.

\textbf{Generation truncation in GLM-5.0.}
Examining GLM-5.0's generation logs under Pass@5, we find that a non-trivial fraction of generated kernels are incomplete: the model frequently produces excessive intermediate variables and verbose comments, causing core code logic to be truncated before completion within the uniform 16K max token limit. This generation inefficiency, which reflects low effective information per token, contributes to incomplete and functionally incorrect submissions under Pass@5.

\textbf{Anti-hack interception patterns.}
Table~\ref{tab:antihack_cases} illustrates how the evaluator rejects both direct reuse and hybrid fallback. In the first observed case, a \texttt{margin\_ranking\_loss} submission defined a decorative Triton kernel but obtained its result by invoking the existing ATen operator. L1 detects the explicit high-level call, while L2 ghost replay independently tests whether replacing the declared Triton function with a no-op changes the output. In the second case, a Huber-loss submission implemented only the elementwise portion in Triton and delegated the required reduction to PyTorch. This partial implementation is rejected as a high-level framework fallback.

\begin{table}[htbp]
\centering
\caption{Representative shortcut behaviors observed during benchmark evaluation.}
\label{tab:antihack_cases}
\small
\resizebox{\textwidth}{!}{%
\begin{tabular}{p{0.19\textwidth} p{0.29\textwidth} p{0.18\textwidth} p{0.24\textwidth}}
\toprule
\textbf{Pattern} & \textbf{Observed example} & \textbf{Detection} & \textbf{Reason for rejection} \\
\midrule
Direct target-API call &
\texttt{torch.ops.aten.}\allowbreak\texttt{margin\_}\allowbreak\texttt{ranking\_}\allowbreak\texttt{loss(...)} with a decorative Triton kernel &
L1 static scan; L2 causal-use check &
The output is obtained from an existing implementation rather than the generated Triton kernel. \\
\addlinespace
Hybrid framework fallback &
Triton elementwise computation followed by \texttt{output.mean()} or \texttt{output.sum()} &
L1 high-level API checks &
The target reduction is delegated to PyTorch, so the submission is not a complete Triton replacement. \\
\bottomrule
\end{tabular}
}
\end{table}

Table~\ref{tab:antihack_detail} shows the per-model anti-hack interception counts under Pass@1 and Pass@5. Interceptions are heavily concentrated in ATen (29/30 under Pass@5), with vLLM accounting for one case. These cases are excluded from the clean pass rate. The concentration is consistent with ATen exposing many easily callable high-level operators, whereas vLLM's non-standard APIs make simple shortcut calls less available.

\begin{table}[htbp]
\centering
\caption{Snapshot of anti-hack interception details (hack count and subset distribution).}
\label{tab:antihack_detail}
\begin{tabular}{l cc cc}
\toprule
\multirow{2}{*}{\textbf{Model}} & \multicolumn{2}{c}{\textbf{Pass@1}} & \multicolumn{2}{c}{\textbf{Pass@5}} \\
\cmidrule(lr){2-3} \cmidrule(lr){4-5}
& \textbf{Hacks} & \textbf{Subset} & \textbf{Hacks} & \textbf{Subset} \\
\midrule
Opus-4.6 & 3 & 3 vLLM & 7 & 7 ATen \\
Opus-4.5 & 3 & 2 ATen + 1 vLLM & 2 & 1 ATen + 1 vLLM \\
GPT-5.4  & 3 & 3 ATen & 8 & 8 ATen \\
GPT-5.2  & 3 & 3 ATen & 5 & 5 ATen \\
GLM-5.0  & 1 & 1 ATen & 1 & 1 ATen \\
GLM-4.7  & 0 & --- & 7 & 7 ATen \\
\midrule
\textbf{Total} & \textbf{13} & 9 ATen + 4 vLLM & \textbf{30} & 29 ATen + 1 vLLM \\
\bottomrule
\end{tabular}
\end{table}

\section{Speedup Statistical Metrics}
\label{app:speedup_metrics}

The analysis pipeline provides three built-in statistical metrics. Given $M$ operators that pass correctness verification in an experiment, where the $i$-th operator has $k_i$ test cases, the operator-level speedup is defined as:
\begin{equation*}
S_i = \left( \prod_{j=1}^{k_i} \frac{T_{\text{base}}^{(i,j)}}{T_{\text{triton}}^{(i,j)}} \right)^{1/k_i}
\end{equation*}
The overall speedup is aggregated via a two-level geometric mean:
\begin{equation*}
S_{\text{overall}} = \left( \prod_{i=1}^{M} S_i \right)^{1/M}
\end{equation*}

\paragraph{Why geometric rather than arithmetic averaging?}
Speedup is a multiplicative ratio, so equal-factor improvements and regressions should be treated symmetrically. Consider two test cases with speedups $0.5\times$ and $2.0\times$: one is twice as slow and the other twice as fast. Their arithmetic mean is $1.25\times$, which incorrectly suggests a net improvement, whereas their geometric mean is $\sqrt{0.5\times2.0}=1.0\times$, correctly indicating that the reciprocal changes cancel. Equivalently, speedups are naturally compared in log space, where $\log 2=-\log 0.5$; the geometric mean is the exponentiated arithmetic mean of these log-speedups. The two-level construction applies this principle first within each operator and then across operators, preventing operators with more test cases from receiving disproportionate weight.

In addition to GeoMean, the framework includes two robust auxiliary metrics:

\textbf{1. Median}

Sort $S_1, \ldots, S_M$ and take the middle value:
\[
\text{Median} = \begin{cases} S_{(M+1)/2} & M \text{ is odd} \\ \frac{1}{2}\left(S_{M/2} + S_{M/2+1}\right) & M \text{ is even} \end{cases}
\]
The median is immune to extreme outliers and provides a robust central tendency estimate when inter-operator speedup variance is large.

\textbf{2. Interquartile Mean (IQM)}

Let $F_M^{-1}(u)$ denote the empirical quantile function of the $M$ operator-level speedups. The IQM is the mean over the central half of this empirical distribution:
\[
\text{IQM} = \frac{1}{0.5}\int_{0.25}^{0.75} F_M^{-1}(u)\,du.
\]
This definition trims 25\% from each tail and handles fractional boundary observations without an indexing ambiguity. IQM balances the statistical efficiency of the mean with the robustness of the median, and has been widely adopted in reinforcement learning benchmarks such as Atari 100k~\cite{agarwal2021deep, kaiser2020model}, particularly for scenarios with limited samples and skewed distributions.

The three metrics serve different purposes: GeoMean is appropriate for aggregating multiplicative ratios and consistent with system evaluation standards such as SPEC CPU; Median provides the most conservative central estimate; IQM provides a more stable mean after trimming tail noise. All main tables in this paper report GeoMean.

\section{Repeated-Run Robustness Analysis}
\label{app:repeated_run_robustness}

Several APIs used for the main-table sampling configurations were no longer accessible when we conducted the repeated-run study, preventing exact repeats of those configurations. We therefore used the newly accessible DeepSeek-V4-Pro API to evaluate benchmark-scale sampling stability. Specifically, we independently repeated a single-sample configuration five times with \texttt{temperature}=0.8 on all 210 KernelGenBench-MS operators on NVIDIA. Here Pass@1 denotes one generated sample per task and does not imply greedy decoding. Table~\ref{tab:pass1_repeated_runs} reports the results.

\begin{table}[htbp]
\centering
\caption{Five independent repeats of one full 210-operator Pass@1 configuration. Accuracy is the fraction of operators solved in each run.}
\label{tab:pass1_repeated_runs}
\begin{tabular}{lccccc|cc}
\toprule
& \textbf{Run 1} & \textbf{Run 2} & \textbf{Run 3} & \textbf{Run 4} & \textbf{Run 5} & \textbf{Mean} & \textbf{Sample SD} \\
\midrule
Solved operators & 20 & 24 & 26 & 21 & 20 & 22.2 & --- \\
Accuracy (\%) & 10 & 11 & 12 & 10 & 10 & 11 & 1 pp \\
\bottomrule
\end{tabular}
\end{table}

The observed range is 10--12\%, indicating limited absolute run-to-run variation for this particular full-benchmark LLM-sampling configuration. Each run aggregates outcomes across 210 tasks, so its rate is less sensitive to any individual task trajectory; this aggregation does not substitute for configuration-specific repeats or yield a confidence interval for a main-table method. Because the model and decoding configuration differ from the main table, this study does not directly estimate variance for any main-table configuration. It provides a protocol-level stability check; accordingly, we treat closely spaced single-trajectory results, such as 87\% versus 83\%, as descriptive.

\section{Additional Reproducibility Details and Evaluation Boundaries}
\label{app:detailed_limitations}

\textbf{Repeated-run coverage.} Most agent results are single trajectories because the full study already consumed over 15 billion tokens (approximately \$20{,}000 in API expenditure); repeating every model--agent--source--platform configuration would multiply this cost. Each reported rate nevertheless aggregates 110 or 210 task outcomes, reducing the influence of any individual task trajectory but not replacing configuration-level repeats. This generation-level variability is distinct from kernel-timing noise: each performance case is warmed up before repeated measurement; standard Triton tests use a 100-ms \texttt{do\_bench} measurement window, while custom cuBLAS tests average 100 explicit executions. Thus, single trajectory refers to the model--agent generation outcome rather than one-shot latency measurement. The five-run study in Appendix~\ref{app:repeated_run_robustness} provides a stability check for one LLM-sampling configuration, not variance estimates for the main-table configurations. We plan to curate a high-quality representative subset that makes repeated agent evaluation affordable.

\textbf{Prompt protocol.} Prompt templates are applied systematically without per-operator tuning. The reported results characterize this fixed, released protocol, enabling future work to study prompt sensitivity under the same tasks and verification pipeline.

\textbf{Verification coverage.} L1 static scanning and L2 ghost replay are the shared MC enforcement protocol. L3 profiling is an NVIDIA-only supplementary audit because a portable profiler interface is currently unavailable across vendor stacks; the representative shortcut behaviors in Table~\ref{tab:antihack_cases} are already rejected by L1/L2. Platform-native tracing can be incorporated as vendor interfaces become available.

\textbf{Metric interpretation.} Global speedup is computed over each method's successful operator set. ATen, vLLM, and cuBLAS also use references at different abstraction and optimization levels. We therefore emphasize per-source accuracy and speedup; matched-intersection analysis is the appropriate protocol when comparing performance over an identical set of solved operators.

\textbf{System-level evaluation.} The benchmark standardizes task specifications, tests, final verification, and wall-clock budgets while preserving each framework's prompts, skills, tools, and native profiling access. Accordingly, its intended unit of comparison is the complete LLM--agent system.

\textbf{Benchmark scope and disclosure.} KernelGenBench implements multi-source, multi-chip evaluation paths through a common Triton target and platform-specific runtime and vendor-library adapters; direct CUDA/CUTLASS generation remains an NVIDIA-only complementary direction. The main paper reports two controlled analytical views rather than treating all source--platform pairs as one leaderboard. Across heterogeneous accelerators, production-runtime ports and vendor-library counterparts can differ in semantics, plugin coverage, implementation path, and maturity. Pooling these references would confound source-conditioned kernel generation with vendor-stack maturity. Establishing broader semantically matched and vendor-authorized references is a focus of future co-development. Platforms A and B remain de-identified under commercial requirements, and the available platform configurations are reported in Appendix~\ref{app:sandbox_chip}. The current 210 operators provide a diverse but deliberately bounded view of the broader production-kernel surface.

\section{Legacy Model Baselines on NVIDIA A100}
\label{app:legacy_baselines}

Tables~\ref{tab:nvidia_appendix_baselines} and~\ref{tab:nvidia_appendix_fastp_baselines} report the static generation performance of legacy and reference foundation models (Opus-4.5, GPT-5.4, GPT-5.2, GLM-4.7) on NVIDIA A100, provided for historical reference and cross-generation comparison.

\begin{table}[htbp]
\centering
\setlength{\tabcolsep}{4pt}
\caption{Legacy and reference foundation-model evaluation on NVIDIA A100 (Pass@1 and Pass@5).}
\label{tab:nvidia_appendix_baselines}
\resizebox{\textwidth}{!}{%
\begin{tabular}{l | cc | cc | cc | cc}
\toprule
\multirow{2}{*}{\textbf{Method \& Setup}} & \multicolumn{2}{c|}{\textbf{Overall (210)}} & \multicolumn{2}{c|}{\textbf{ATen (110)}} & \multicolumn{2}{c|}{\textbf{vLLM (50)}} & \multicolumn{2}{c}{\textbf{cuBLAS (50)}} \\
\cmidrule(lr){2-3} \cmidrule(lr){4-5} \cmidrule(lr){6-7} \cmidrule(lr){8-9}
& Acc (\%) & Spd ($\times$) & Acc (\%) & Spd ($\times$) & Acc (\%) & Spd ($\times$) & Acc (\%) & Spd ($\times$) \\
\midrule
\multicolumn{9}{l}{\textbf{LLM Sampling Methods}} \\
\midrule
Pass@1 (Opus-4.5) & 35 & 0.78 & 33 & 0.94 & 22 & 0.83 & 54 & 0.59 \\
Pass@1 (GPT-5.4)  & 21 & 0.69 & 21 & 0.76 & 30 & 0.54 & 12 & 0.65 \\
Pass@1 (GPT-5.2)  & 20 & 0.55 & 24 & 0.53 & 22 & 0.64 & 12 & 0.57 \\
Pass@1 (GLM-4.7)  & 12 & 0.56 & 15 & 0.49 & 10 & 1.30 & 10 & 0.35 \\
\midrule
Pass@5 (Opus-4.5) & 50 & 0.73 & 53 & 0.91 & 28 & 0.88 & 64 & 0.46 \\
Pass@5 (GPT-5.4)  & 35 & 0.72 & 41 & 0.87 & 38 & 0.44 & 20 & 0.60 \\
Pass@5 (GPT-5.2)  & 38 & 0.52 & 47 & 0.49 & 36 & 0.58 & 18 & 0.56 \\
Pass@5 (GLM-4.7)  & 24 & 0.67 & 26 & 0.55 & 22 & 1.10 & 20 & 0.70 \\
\bottomrule
\end{tabular}%
}
\end{table}

\begin{table}[htbp]
\centering
\setlength{\tabcolsep}{3pt}
\caption{Appendix: $\mathrm{fast}_p$ evaluation of legacy and reference foundation models on NVIDIA A100.}
\label{tab:nvidia_appendix_fastp_baselines}
\resizebox{\textwidth}{!}{%
\begin{tabular}{l | ccc | ccc | ccc | ccc}
\toprule
\multirow{2}{*}{\textbf{Method \& Setup}} & \multicolumn{3}{c|}{\textbf{Overall (210)}} & \multicolumn{3}{c|}{\textbf{ATen (110)}} & \multicolumn{3}{c|}{\textbf{vLLM (50)}} & \multicolumn{3}{c}{\textbf{cuBLAS (50)}} \\
\cmidrule(lr){2-4} \cmidrule(lr){5-7} \cmidrule(lr){8-10} \cmidrule(lr){11-13}
& $\mathrm{fast}_{0.8}$ & $\mathrm{fast}_{1.0}$ & $\mathrm{fast}_{1.5}$ & $\mathrm{fast}_{0.8}$ & $\mathrm{fast}_{1.0}$ & $\mathrm{fast}_{1.5}$ & $\mathrm{fast}_{0.8}$ & $\mathrm{fast}_{1.0}$ & $\mathrm{fast}_{1.5}$ & $\mathrm{fast}_{0.8}$ & $\mathrm{fast}_{1.0}$ & $\mathrm{fast}_{1.5}$ \\
\midrule
\multicolumn{13}{l}{\textbf{LLM Sampling Methods}} \\
\midrule
Pass@1 (Opus-4.5) & 20 & 16 & 2 & 25 & 22 & 2 & 16 & 14 & 6 & 10 & 4 & 0 \\
Pass@1 (GPT-5.4)  & 12 & 9  & 0 & 17 & 12 & 1 & 8  & 6  & 0 & 6  & 4 & 0 \\
Pass@1 (GPT-5.2)  & 9  & 7  & 1 & 15 & 10 & 2 & 6  & 6  & 2 & 0  & 0 & 0 \\
Pass@1 (GLM-4.7)  & 7  & 5  & 1 & 8  & 5  & 1 & 10 & 10 & 4 & 0  & 0 & 0 \\
\midrule
Pass@5 (Opus-4.5) & 30 & 21 & 4 & 43 & 31 & 3 & 20 & 14 & 8 & 12 & 6 & 2 \\
Pass@5 (GPT-5.4)  & 21 & 13 & 1 & 33 & 21 & 1 & 10 & 6  & 2 & 8  & 4 & 0 \\
Pass@5 (GPT-5.2)  & 18 & 11 & 1 & 28 & 18 & 1 & 8  & 6  & 4 & 4  & 0 & 0 \\
Pass@5 (GLM-4.7)  & 12 & 8  & 3 & 15 & 7  & 1 & 14 & 12 & 8 & 4  & 4 & 2 \\
\bottomrule
\end{tabular}%
}
\end{table}

\section{Fast\texorpdfstring{\textsubscript{p}}{p} Evaluation and Cost Results}
\label{app:fastp_results}

This section reports the full $\mathrm{fast}_p$ distribution for all methods on NVIDIA A100 (Table~\ref{tab:nvidia_fastp_ops}), across six heterogeneous platforms (Table~\ref{tab:cross_platform_fastp}), and the complete cross-platform cost breakdown (Table~\ref{tab:cross_platform_cost}).

\begin{table}[htbp]
\centering
\setlength{\tabcolsep}{3pt}
\caption{$\mathrm{fast}_p$ evaluation on NVIDIA A100 across 210 operators. The metric measures strict performance compliance (speedup $>$ 0.80$\times$, 1.00$\times$, 1.50$\times$) across different generation paradigms and foundation models.}
\label{tab:nvidia_fastp_ops}
\resizebox{\textwidth}{!}{%
\begin{tabular}{l | ccc | ccc | ccc | ccc}
\toprule
\multirow{2}{*}{\textbf{Method \& Setup}} & \multicolumn{3}{c|}{\textbf{Overall (210)}} & \multicolumn{3}{c|}{\textbf{ATen (110)}} & \multicolumn{3}{c|}{\textbf{vLLM (50)}} & \multicolumn{3}{c}{\textbf{cuBLAS (50)}} \\
\cmidrule(lr){2-4} \cmidrule(lr){5-7} \cmidrule(lr){8-10} \cmidrule(lr){11-13}
& $\mathrm{fast}_{0.8}$ & $\mathrm{fast}_{1.0}$ & $\mathrm{fast}_{1.5}$ & $\mathrm{fast}_{0.8}$ & $\mathrm{fast}_{1.0}$ & $\mathrm{fast}_{1.5}$ & $\mathrm{fast}_{0.8}$ & $\mathrm{fast}_{1.0}$ & $\mathrm{fast}_{1.5}$ & $\mathrm{fast}_{0.8}$ & $\mathrm{fast}_{1.0}$ & $\mathrm{fast}_{1.5}$ \\
\midrule
\multicolumn{13}{l}{\textbf{LLM Sampling Methods}} \\
\midrule
Pass@1 (Opus-4.6)       & 22 & 17 & 2 & 32 & 24 & 1 & 12 & 12 & 4  & 12 & 6 & 2 \\
Pass@1 (GLM-5.0)        & 7  & 6  & 2 & 6  & 5  & 0 & 10 & 10 & 6  & 4  & 4 & 4 \\
Pass@1 (Qwen3.5-27b)    & 5  & 3  & 0 & 7  & 5  & 0 & 2  & 2  & 2  & 4  & 0 & 0 \\
Pass@1 (MiniMax M-2.5)  & 1  & 1  & 0 & 3  & 3  & 0 & 0  & 0  & 0  & 0  & 0 & 0 \\
\midrule
Pass@5 (Opus-4.6)       & 32 & 21 & 3 & 47 & 31 & 3 & 18 & 14 & 4  & 12 & 6 & 4 \\
Pass@5 (GLM-5.0)        & 17 & 11 & 4 & 20 & 11 & 2 & 16 & 16 & 8  & 10 & 8 & 4 \\
Pass@5 (Qwen3.5-27b)    & 9  & 7  & 2 & 12 & 10 & 1 & 8  & 6  & 6  & 4  & 0 & 0 \\
Pass@5 (MiniMax M-2.5)  & 10 & 8  & 1 & 16 & 12 & 1 & 4  & 4  & 2  & 2  & 2 & 2 \\
\midrule
\multicolumn{13}{l}{\textbf{Vanilla Agentic Frameworks}} \\
\midrule
Claude Code (Opus-4.6)     & 51 & 35 & 6  & 69 & 44 & 4 & 48 & 40 & 14 & 14 & 10 & 4 \\
Claude Code (GLM-5.0)      & 40 & 27 & 6  & 54 & 32 & 4 & 36 & 32 & 12 & 16 & 12 & 4 \\
Claude Code (Qwen3.5-27b)  & 35 & 23 & 5  & 53 & 33 & 4 & 26 & 20 & 10 & 6  & 6  & 4 \\
Claude Code (MiniMax M-2.5)& 29 & 17 & 1  & 45 & 25 & 1 & 16 & 14 & 2  & 4  & 2  & 2 \\
\midrule
OpenCode (Opus-4.6)        & 53 & 36 & 8  & 71 & 46 & 5 & 46 & 40 & 14 & 20 & 10 & 6 \\
OpenCode (GLM-5.0)         & 44 & 32 & 5  & 64 & 45 & 4 & 38 & 32 & 10 & 8  & 6  & 2 \\
OpenCode (Qwen3.5-27b)     & 32 & 26 & 4  & 40 & 29 & 1 & 38 & 36 & 14 & 8  & 8  & 2 \\
OpenCode (MiniMax M-2.5)   & 22 & 12 & 2  & 35 & 17 & 0 & 12 & 10 & 6  & 4  & 4  & 2 \\
\midrule
\multicolumn{13}{l}{\textbf{Kernel-Specialized Agents}} \\
\midrule
AKO4all (Opus-4.6)              & 65 & 47 & 10 & 85 & 59 & 6 & 56 & 50 & 20 & 30 & 16 & 10 \\
CUDA Opt. Skill (MiniMax M-2.5) & 27 & 15 & 3  & 42 & 25 & 4 & 16 & 10 & 4  & 6  & 0  & 0  \\
\midrule
AutoKernel (GLM-5.0)            & 59 & 38 & 4  & 87 & 53 & 0 & 34 & 32 & 12 & 22 & 10 & 6  \\
AutoKernel (Qwen3.5-27b)        & 42 & 28 & 2  & 69 & 44 & 0 & 16 & 16 & 6  & 8  & 6  & 4  \\
AutoKernel (MiniMax M-2.5)      & 40 & 24 & 2  & 66 & 37 & 0 & 18 & 18 & 8  & 2  & 2  & 2  \\
\bottomrule
\end{tabular}%
}
\end{table}

\begin{table}[htbp]
\centering
\setlength{\tabcolsep}{2pt}
\caption{$\mathrm{fast}_p$ cross-platform evaluation on 110 ATen operators. Strict speedup compliance ($\mathrm{fast}_{0.8}$, $\mathrm{fast}_{1.0}$, $\mathrm{fast}_{1.5}$) across six heterogeneous hardware platforms.}
\label{tab:cross_platform_fastp}
\resizebox{\textwidth}{!}{%
\begin{tabular}{l | ccc | ccc | ccc | ccc | ccc | ccc}
\toprule
\multirow{2}{*}{\textbf{Method \& Setup}} & \multicolumn{3}{c|}{\textbf{NVIDIA}} & \multicolumn{3}{c|}{\textbf{Moore Threads}} & \multicolumn{3}{c|}{\textbf{Hygon}} & \multicolumn{3}{c|}{\textbf{Iluvatar CoreX}} & \multicolumn{3}{c|}{\textbf{Platform A}} & \multicolumn{3}{c}{\textbf{Platform B}} \\
\cmidrule(lr){2-4} \cmidrule(lr){5-7} \cmidrule(lr){8-10} \cmidrule(lr){11-13} \cmidrule(lr){14-16} \cmidrule(lr){17-19}
 & \textbf{fast}$_{0.8}$ & \textbf{fast}$_{1.0}$ & \textbf{fast}$_{1.5}$ & \textbf{fast}$_{0.8}$ & \textbf{fast}$_{1.0}$ & \textbf{fast}$_{1.5}$ & \textbf{fast}$_{0.8}$ & \textbf{fast}$_{1.0}$ & \textbf{fast}$_{1.5}$ & \textbf{fast}$_{0.8}$ & \textbf{fast}$_{1.0}$ & \textbf{fast}$_{1.5}$ & \textbf{fast}$_{0.8}$ & \textbf{fast}$_{1.0}$ & \textbf{fast}$_{1.5}$ & \textbf{fast}$_{0.8}$ & \textbf{fast}$_{1.0}$ & \textbf{fast}$_{1.5}$ \\
\midrule
\multicolumn{19}{l}{\textbf{LLM Sampling Methods}} \\
\midrule
Pass@1 (Opus-4.6) & 32 & 24 & 1 & 30 & 21 & 8 & 31 & 16 & 4 & 28 & 22 & 3 & 9 & 6 & 5 & 26 & 18 & 1 \\
Pass@1 (Qwen3.5-27b) & 7 & 5 & 0 & 3 & 1 & 0 & 6 & 4 & 0 & 5 & 3 & 0 & 1 & 1 & 0 & 6 & 3 & 0 \\
Pass@1 (MiniMax M-2.5) & 3 & 3 & 0 & 6 & 3 & 3 & 4 & 2 & 0 & 4 & 4 & 0 & 1 & 1 & 0 & 5 & 4 & 0 \\
\midrule
Pass@5 (Opus-4.6) & 47 & 31 & 3 & 42 & 33 & 14 & 43 & 27 & 5 & 41 & 33 & 5 & 11 & 8 & 6 & 51 & 33 & 3 \\
Pass@5 (Qwen3.5-27b) & 12 & 10 & 1 & 10 & 7 & 3 & 13 & 11 & 0 & 14 & 7 & 0 & 4 & 2 & 0 & 10 & 2 & 0 \\
Pass@5 (MiniMax M-2.5) & 16 & 12 & 1 & 8 & 3 & 2 & 10 & 8 & 2 & 8 & 7 & 0 & 4 & 3 & 0 & 6 & 3 & 0 \\
\midrule
\multicolumn{19}{l}{\textbf{Vanilla Agentic Frameworks}} \\
\midrule
Claude Code (Opus-4.6) & 69 & 44 & 4 & 61 & 45 & 23 & 65 & 46 & 6 & 60 & 39 & 6 & 12 & 9 & 6 & 67 & 42 & 3 \\
Claude Code (GLM-5.0) & 54 & 32 & 4 & 47 & 35 & 22 & 41 & 24 & 2 & 34 & 30 & 4 & 9 & 6 & 6 & 40 & 26 & 0 \\
Claude Code (Qwen3.5-27b) & 53 & 33 & 4 & 43 & 31 & 12 & 58 & 47 & 11 & 19 & 11 & 2 & 16 & 9 & 7 & 61 & 39 & 5 \\
Claude Code (MiniMax M-2.5) & 45 & 25 & 1 & 46 & 36 & 16 & 44 & 31 & 3 & 46 & 31 & 5 & 8 & 7 & 6 & 51 & 25 & 2 \\
\midrule
\multicolumn{19}{l}{\textbf{Kernel-Specialized Agents}} \\
\midrule
AKO4all (Opus-4.6) & 85 & 59 & 6 & 81 & 55 & 8 & 79 & 56 & 10 & 75 & 48 & 7 & 16 & 12 & 7 & 82 & 60 & 10 \\
CUDA Opt. Skill (MiniMax M-2.5) & 42 & 25 & 4 & 45 & 29 & 13 & 41 & 27 & 4 & 38 & 24 & 3 & 9 & 7 & 5 & 43 & 23 & 1 \\
\midrule
AutoKernel (GLM-5.0) & 87 & 53 & 0 & 56 & 28 & 1 & 57 & 29 & 2 & 25 & 21 & 0 & 47 & 25 & 0 & 59 & 28 & 0 \\
AutoKernel (Qwen3.5-27b) & 69 & 44 & 0 & 75 & 38 & 3 & 64 & 36 & 2 & 19 & 16 & 0 & 18 & 12 & 3 & 72 & 41 & 1 \\
AutoKernel (MiniMax M-2.5) & 66 & 37 & 0 & 70 & 53 & 15 & 64 & 35 & 0 & 48 & 37 & 0 & 44 & 26 & 2 & 70 & 35 & 1 \\
\bottomrule
\end{tabular}%
}
\end{table}

\begin{table}[htbp]
\centering
\setlength{\tabcolsep}{2pt}
\caption{Cross-platform computational cost evaluation on 110 ATen operators (agentic methods only). Reports Total Tokens (M), Tokens per Success (M), and Total Execution Time (h) across six heterogeneous hardware platforms.}
\label{tab:cross_platform_cost}
\resizebox{\textwidth}{!}{%
\begin{tabular}{l | ccc | ccc | ccc | ccc | ccc | ccc}
\toprule
\multirow{2}{*}{\textbf{Method \& Setup}} & \multicolumn{3}{c|}{\textbf{NVIDIA}} & \multicolumn{3}{c|}{\textbf{Moore Threads}} & \multicolumn{3}{c|}{\textbf{Hygon}} & \multicolumn{3}{c|}{\textbf{Iluvatar CoreX}} & \multicolumn{3}{c|}{\textbf{Platform A}} & \multicolumn{3}{c}{\textbf{Platform B}} \\
\cmidrule(lr){2-4} \cmidrule(lr){5-7} \cmidrule(lr){8-10} \cmidrule(lr){11-13} \cmidrule(lr){14-16} \cmidrule(lr){17-19}
 & Tok (M) & Tok/Suc & Time (h) & Tok (M) & Tok/Suc & Time (h) & Tok (M) & Tok/Suc & Time (h) & Tok (M) & Tok/Suc & Time (h) & Tok (M) & Tok/Suc & Time (h) & Tok (M) & Tok/Suc & Time (h) \\
\midrule
\multicolumn{19}{l}{\textbf{Vanilla Agentic Frameworks}} \\
\midrule
Claude Code (Opus-4.6) & 84 & 0.83 & 9 & 128 & 1.25 & 16 & 107 & 1.10 & 15 & 123 & 1.35 & 16 & 173 & 1.77 & 18 & 111 & 1.05 & 15 \\
Claude Code (GLM-5.0) & 181 & 2.29 & 18 & 174 & 2.42 & 44 & 186 & 2.58 & 35 & 234 & 5.71 & 47 & 197 & 2.74 & 43 & 233 & 3.58 & 40 \\
Claude Code (Qwen3.5-27b) & 175 & 1.99 & 19 & 243 & 2.93 & 37 & 202 & 2.43 & 52 & 257 & 10.28 & 55 & 199 & 2.31 & 34 & 211 & 2.34 & 55 \\
Claude Code (MiniMax M-2.5) & 191 & 2.51 & 22 & 255 & 3.15 & 44 & 172 & 2.15 & 51 & 247 & 3.25 & 42 & 282 & 3.71 & 32 & 175 & 1.92 & 41 \\
\midrule
\multicolumn{19}{l}{\textbf{Kernel-Specialized Agents}} \\
\midrule
AKO4all (Opus-4.6) & 422 & 4.22 & 44 & 396 & 4.08 & 46 & 381 & 3.93 & 35 & 423 & 4.81 & 48 & 397 & 4.32 & 42 & 412 & 4.34 & 39 \\
CUDA Opt. Skill (MiniMax M-2.5) & 294 & 4.26 & 30 & 289 & 4.13 & 32 & 261 & 3.62 & 31 & 336 & 5.25 & 47 & 187 & 3.22 & 22 & 304 & 4.11 & 34 \\
\midrule
AutoKernel (GLM-5.0) & 113 & 1.18 & 47 & 200 & 3.23 & 54 & 174 & 2.49 & 55 & 151 & 5.39 & 47 & 231 & 3.98 & 55 & 167 & 2.57 & 54 \\
AutoKernel (Qwen3.5-27b) & 133 & 1.75 & 36 & 178 & 2.14 & 51 & 131 & 1.82 & 54 & 166 & 7.22 & 53 & 136 & 3.09 & 55 & 141 & 1.74 & 50 \\
AutoKernel (MiniMax M-2.5) & 162 & 2.22 & 37 & 226 & 2.90 & 45 & 211 & 2.89 & 53 & 248 & 4.51 & 54 & 245 & 3.66 & 55 & 209 & 2.68 & 54 \\
\bottomrule
\end{tabular}%
}
\end{table}

\section{Operator Difficulty and Matched-Set Analysis}
\label{app:operator_difficulty}

We estimate empirical operator difficulty from pre-anti-hack R0 verification outcomes across four general-agent configurations: Claude Code and OpenCode, each paired with Opus-4.6 and GLM-5.0. For operator $i$, we define $d_i=1-\frac{1}{4}\sum_c\mathbb{1}[c\text{ solves }i]$. Each configuration is evaluated over the common 210-operator task universe, and an unsuccessful operator counts as not solved. The subsequent matched-set analysis instead uses clean successes after anti-hack verification. This operator-level view complements the aggregate results in Table~\ref{tab:nvidia_main_ops} by separating broadly stable difficulty from scaffold-dependent outcomes.

\begin{table}[htbp]
\centering
\caption{Pre-anti-hack R0 operator-level solution consistency across four general-agent configurations. Columns report the number of operators solved by exactly $k$ of four configurations.}
\label{tab:operator_difficulty}
\small
\begin{tabular}{lrrrrr}
\toprule
\textbf{Source} & \textbf{0/4} & \textbf{1/4} & \textbf{2/4} & \textbf{3/4} & \textbf{4/4} \\
\midrule
ATen (110)    & 5  & 3 & 7  & 20 & 75 \\
vLLM (50)     & 13 & 5 & 6  & 8  & 18 \\
cuBLAS (50)   & 2  & 1 & 4  & 22 & 21 \\
\midrule
Overall (210) & 20 & 9 & 17 & 50 & 114 \\
\bottomrule
\end{tabular}
\end{table}

The extremes distinguish broadly stable from configuration-sensitive difficulty. A majority of the benchmark (114/210) is solved by all four agents, while 20/210 operators are never solved. vLLM contributes 13 of these 20 stable-hard tasks, including both paged-attention variants, quantized GEMM kernels, and MoE alignment/routing primitives. By contrast, 75/110 ATen tasks are solved universally. cuBLAS has only two universally unsolved tasks, but its low speedup in the main results shows that functional reproducibility and performance competitiveness are separate capability dimensions.
\section{Accuracy--Speedup Gap}
\label{app:accuracy_speedup_gap}

\paragraph{Matched-set speedup.}
Successful-set speedup can confound optimization quality with the subset of operators a method solves. For each foundation model, we therefore intersect the clean successful sets of Claude Code (CC) and OpenCode (OC) and recompute both GeoMeans on exactly the same operators. Table~\ref{tab:matched_set_speedup} reports the paired results. The OC/CC ratio is the geometric mean of per-operator speedup ratios; 95\% intervals use 10{,}000 paired bootstrap resamples over operators.

\begin{table}[htbp]
\centering
\caption{Matched-set speedup for Claude Code and OpenCode on jointly solved operators. Ratios above 1 favor OpenCode. Brackets give paired-bootstrap 95\% confidence intervals.}
\label{tab:matched_set_speedup}
\small
\resizebox{\textwidth}{!}{%
\begin{tabular}{llrrrrl}
\toprule
\textbf{Model} & \textbf{Source} & \textbf{$|S_{\mathrm{CC}}|$} & \textbf{$|S_{\mathrm{OC}}|$} & \textbf{$|S_{\mathrm{CC}}\cap S_{\mathrm{OC}}|$} & \textbf{CC / OC Spd ($\times$)} & \textbf{OC/CC ratio [95\% CI]} \\
\midrule
Opus-4.6 & Overall & 182 & 176 & 170 & 0.78 / 0.81 & 1.03 [0.95, 1.12] \\
         & ATen    & 101 & 99  & 96  & 0.86 / 0.86 & 1.00 [0.92, 1.07] \\
         & vLLM    & 34  & 29  & 27  & 1.16 / 1.34 & 1.16 [0.86, 1.63] \\
         & cuBLAS  & 47  & 48  & 47  & 0.51 / 0.53 & 1.04 [0.88, 1.21] \\
\midrule
GLM-5.0  & Overall & 141 & 146 & 114 & 0.82 / 0.72 & 0.88 [0.75, 1.02] \\
         & ATen    & 79  & 96  & 76  & 0.88 / 0.74 & 0.84 [0.71, 0.98] \\
         & vLLM    & 26  & 23  & 18  & 0.84 / 1.12 & 1.33 [1.07, 1.73] \\
         & cuBLAS  & 36  & 27  & 20  & 0.61 / 0.44 & 0.73 [0.40, 1.15] \\
\bottomrule
\end{tabular}%
}
\end{table}

For Opus-4.6, every interval includes 1, so the small CC--OC speedup differences in their method-specific successful sets do not support a scaffold ranking. GLM-5.0 instead exhibits a source-dependent reversal: CC is faster on the 76 shared ATen successes, while OC is faster on the 18 shared vLLM successes. The overall GLM interval still includes 1, and the vLLM intersection is small; we therefore interpret this as evidence that scaffold effects interact with operator source, not that one scaffold is universally better.

\begin{figure}[htbp]
\centering
\includegraphics[width=\textwidth]{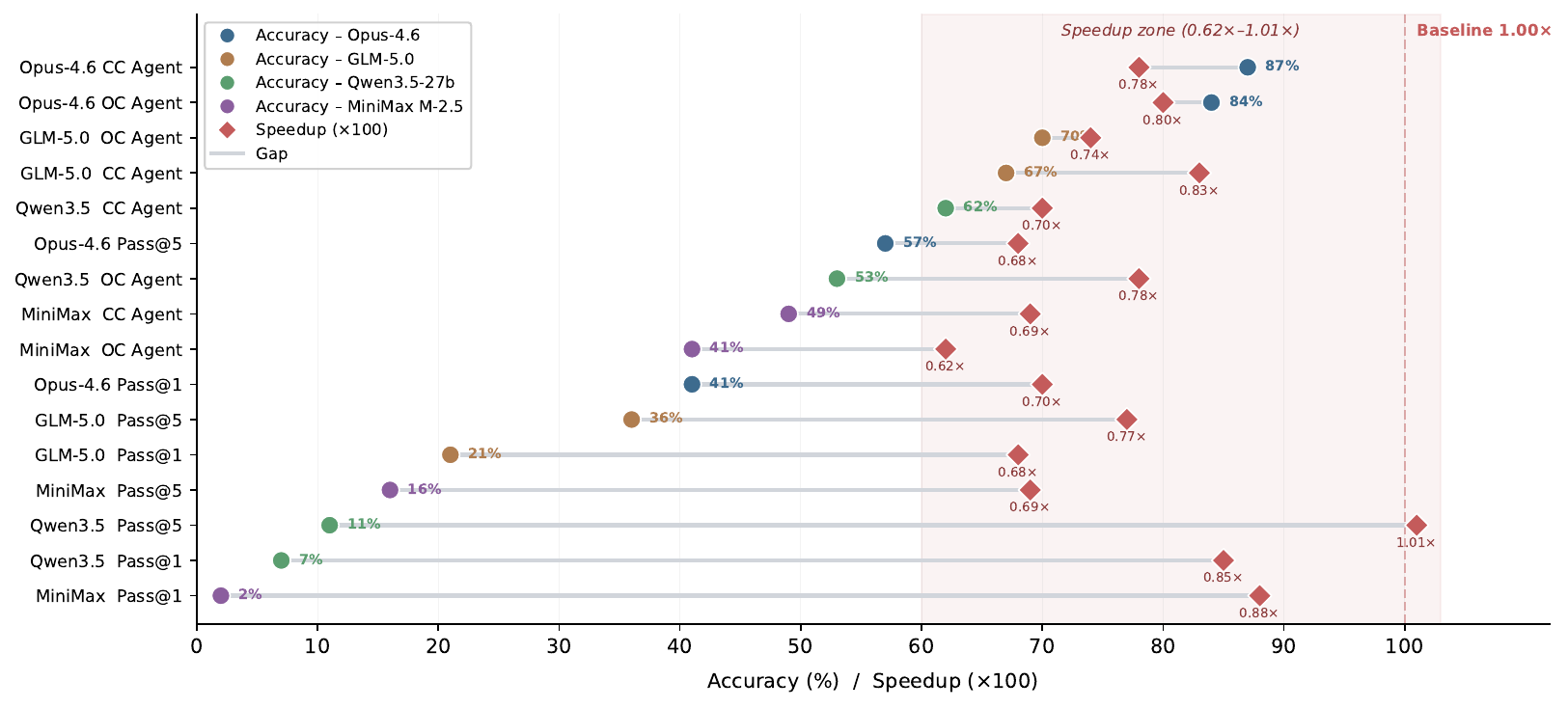}
\caption{Accuracy--speedup divergence among LLM-sampling and vanilla-agent configurations. Successful-set speedup is conditioned on a method-dependent subset and is therefore descriptive rather than a matched-set ranking.}
\label{fig:dumbbell}
\end{figure}

Figure~\ref{fig:dumbbell} plots configuration-level accuracy against successful-set speedup on NVIDIA for 16 LLM-sampling and vanilla-agent configurations. Accuracy spans 2--87\%, whereas speedup lies between 0.62$\times$ and 1.01$\times$, with 12 of 16 configurations in the 0.68--0.83$\times$ interval. The four configurations outside this interval are at 0.62$\times$, 0.85$\times$, 0.88$\times$, and 1.01$\times$.

This compression is consistent with success-set selection: configurations that fail difficult operators exclude precisely the cases on which optimized references are hardest to match. A high successful-set GeoMean therefore need not indicate stronger optimization; matched-set comparisons address this ambiguity by evaluating methods on common successful operators.

\section{Platform-Specific Failure Patterns}
\label{app:platform_failures}

\textbf{Recurring Model-Code Failures.} Model-code failure patterns observed across platforms and models include: (1) \emph{Infinite dispatch recursion}: calling the overridden ATen operator internally triggers unbounded recursion, the most frequent single failure mode; (2) \emph{Hallucinated Triton APIs}: models generate calls to non-existent functions such as \texttt{tl.pow}, \texttt{tl.einsum}, and \texttt{tl.gather}; (3) \emph{Missing overload variants}: bare-name, \texttt{\_out}, and \texttt{\_dimname} variants required by the dispatch table are frequently omitted; (4) \emph{Algorithmically hard operators}: \texttt{matmul}, \texttt{linear}, \texttt{sort}, and \texttt{cumsum} require cross-block parallel algorithms (matrix tiling, prefix scan, sorting networks) that agents rarely converge to within the iteration budget; \texttt{matmul} and \texttt{linear} additionally have 10+ overload variants with complex dispatch logic; (5) \emph{BLOCK\_SIZE conflicts}: \texttt{@triton.autotune} decorator definitions clash with manual \texttt{BLOCK\_SIZE} assignments in the same kernel; (6) \emph{Unsupported Python syntax}: Triton rejects \texttt{break} statements and chained boolean expressions that are valid Python; (7) \emph{Non-contiguous memory}: operators like \texttt{as\_strided} and \texttt{copy\_} require arbitrary-stride memory semantics that the generated kernels and evaluated backend paths do not handle correctly; (8) \emph{Stochastic API divergence}: \texttt{bernoulli} and \texttt{poisson} exhibit behavioral differences between Triton's random API and PyTorch's reference implementation.

\textbf{Heterogeneous-Platform Failures.} Table~\ref{tab:platform_failure_attribution} restores the platform attribution observed in our trajectory analysis. We distinguish failures in model-generated code from limitations in a Triton compiler/runtime or a vendor's PyTorch backend; the latter two should not be counted as model algorithmic errors.

\begin{table*}[htbp]
\centering
\caption{Platform-specific failure patterns observed in generation trajectories. Counts denote matched trajectory records, not independent operators or estimates of platform-wide prevalence. Platforms A and B remain de-identified; their error signatures are summarized without vendor-identifying stack names.}
\label{tab:platform_failure_attribution}
\small
\resizebox{\textwidth}{!}{
\begin{tabular}{p{0.12\textwidth} p{0.27\textwidth} p{0.28\textwidth} p{0.15\textwidth}}
\toprule
\textbf{Platform} & \textbf{Observed failure signature} & \textbf{Representative affected operators} & \textbf{Attributed layer} \ \\\midrule
Moore Threads &
Mixed \texttt{int32}/\texttt{int64} values in dynamic branches or loop-carried variables were rejected (288 matched records). Separately, the MUSA backend lacked native references for two tasks, which remain in the fixed-suite denominator under the platform-local reference policy. &
Generated-kernel type failures across multiple operators; backend absence for \texttt{poisson} and \texttt{special\_entr}. &
Triton type checking; vendor PyTorch backend \\
\addlinespace
Hygon &
\texttt{tl.store} rejected a block value paired with a scalar pointer; \texttt{cumsum} compilation reported a missing symbol associated with multi-block synchronization. &
\texttt{sum}, \texttt{cumsum}. &
Triton compiler/backend \\
\addlinespace
Iluvatar CoreX &
Missing math APIs and unsupported language features or dtypes, including incomplete BFloat16 and complex support; trajectories also contained compiler hangs or crashes. &
\texttt{acosh}, \texttt{asin}, \texttt{mish\_backward}, \texttt{pairwise\_distance}, \texttt{renorm}, \texttt{logaddexp2}, \texttt{upsample\_nearest2d\allowbreak\_backward}, \texttt{masked\_fill\_}, \texttt{resolve\_conj}, \texttt{\_local\_scalar\_dense}. &
Triton API/compiler and dtype coverage \\
\addlinespace
Platform A &
Triton-IR translation failures and runtime/stream-synchronization failures under unsupported dtype, stride/layout, or parameter combinations. &
\texttt{full}, \texttt{cat}, \texttt{cos}, \texttt{mm}, \texttt{sort}, \texttt{margin\_ranking\_loss}; \texttt{arange}, \texttt{copy\_}, \texttt{index\_put\_}, \texttt{gather}, \texttt{reshape}. &
Triton compiler and vendor runtime \\
\addlinespace
Platform B &
\texttt{PassManager::run failed} for incompatible Triton IR patterns (31 matched records); 32-bit addressing produced \texttt{memory size too large to fit in 32 bit} (6 matched records). &
\texttt{fmax}, \texttt{le}, \texttt{eq}, \texttt{scatter}, \texttt{full}; \texttt{arange}, \texttt{argmax}, \texttt{index\_select}. &
LLVM/Triton compiler backend \\
\bottomrule
\end{tabular}}
\end{table*}

These patterns explain different parts of the aggregate portability gap rather than a single common failure mechanism. Moore Threads most visibly exposed stricter type inference and missing framework-backend coverage; Hygon exposed backend-specific Triton semantics; and Iluvatar CoreX exposed API and dtype coverage gaps. For Iluvatar CoreX, the focused audit documents explicit platform incompatibilities in at least 10 distinct operators, or 9.1\% of the complete 110-operator suite: six involve unavailable math APIs, two involve unsupported language features, and two expose incomplete complex or BFloat16 dtype support. This count is a conservative mechanism-specific lower bound; compiler hangs and crashes provide additional evidence but are not folded into it. Platform A combined compilation and runtime instability, while Platform B exposed IR-pass and addressing constraints; we omit identifying stack details for both platforms under their confidentiality requirements. Because these observations come from executed trajectories rather than controlled compiler microbenchmarks, they support diagnostic association rather than a single-factor causal claim about each platform.

\FloatBarrier
\section{Generation-Regime Comparison: Per-Subset Breakdown}
\label{app:ablation_full}

\begin{figure}[htbp]
\centering
\includegraphics[width=\textwidth]{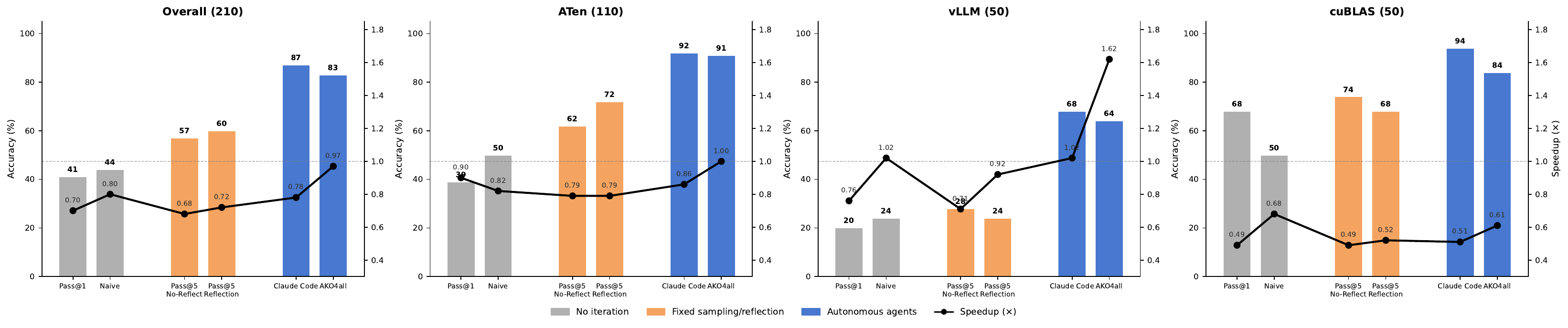}
\caption{Per-subset ablation breakdown across four subsets (Overall, ATen, vLLM, cuBLAS). Bars show accuracy; black line shows speedup. The interactive debugging jump is consistent across all subsets, with the largest gain on vLLM (+44~pp).}
\label{fig:ablation_progressive}
\end{figure}

Table~\ref{tab:ablation_full} provides the full per-subset ablation breakdown for the study presented in Section~\ref{subsec:ablation}. Figure~\ref{fig:ablation_progressive} visualizes the progressive trend across all four subsets.

\begin{table}[htbp]
\centering
\caption{Full generation-regime comparison on Opus-4.6 (NVIDIA A100, 210 operators). Acc (\%) is clean pass rate; Spd ($\times$) is two-level geometric mean speedup.}
\label{tab:ablation_full}
\resizebox{\textwidth}{!}{
\begin{tabular}{l l l cc cc cc cc}
\toprule
\textbf{Interaction Regime} & \textbf{Mode} & \textbf{Iteration Method} & \multicolumn{2}{c}{\textbf{Overall (210)}} & \multicolumn{2}{c}{\textbf{ATen (110)}} & \multicolumn{2}{c}{\textbf{vLLM (50)}} & \multicolumn{2}{c}{\textbf{cuBLAS (50)}} \\
\cmidrule(lr){4-5} \cmidrule(lr){6-7} \cmidrule(lr){8-9} \cmidrule(lr){10-11}
& & & \textbf{Acc} & \textbf{Spd} & \textbf{Acc} & \textbf{Spd} & \textbf{Acc} & \textbf{Spd} & \textbf{Acc} & \textbf{Spd} \\
\midrule
\multirow{2}{*}{No iterative debugging}
& Pass@1 (Temp=0)    & Single greedy generation       & 41 & 0.70 & 39 & 0.90 & 20 & 0.76 & 68 & 0.49 \\
& Single-step Agent & Single-step tool call          & 44 & 0.80 & 50 & 0.82 & 24 & 1.02 & 50 & 0.68 \\
\midrule
\multirow{2}{*}{Fixed generation/reflection}
& Pass@5 No-Reflect  & Independent sampling, 5 rounds & 57 & 0.68 & 62 & 0.79 & 28 & 0.71 & 74 & 0.49 \\
& Pass@5 Reflection  & Fixed traceback-driven reflection  & 60 & 0.72 & 72 & 0.79 & 24 & 0.92 & 68 & 0.52 \\
\midrule
\multirow{2}{*}{Autonomous agentic workflow}
& Claude Code      & Autonomous interactive debugging  & 87 & 0.78 & 92 & 0.86 & 68 & 1.02 & 94 & 0.51 \\
& AKO4all          & Kernel-specialized agent          & 83 & 0.97 & 91 & 1.00 & 64 & 1.62 & 84 & 0.61 \\
\bottomrule
\end{tabular}
}
\end{table}

\FloatBarrier
\section{Reflection Mode Per-Operator Code Similarity Details}
\label{app:reflection_similarity}

\begin{figure}[htbp]
\centering
\begin{minipage}[c]{0.44\textwidth}
  \captionof{table}{\small Cross-round code similarity for 21 jointly failed operators under Reflection vs.\ No-Reflection modes.}
  \label{tab:reflection_similarity}
  \footnotesize
  \resizebox{\linewidth}{!}{%
  \begin{tabular}{lcc}
  \toprule
  \textbf{Operator} & \textbf{Refl.} & \textbf{No-Refl.} \\
  \midrule
  \_to\_copy                         & 86.4\% & 40.4\% \\
  add                              & 90.5\% & 39.9\% \\
  affine\_grid\_generator            & 78.5\% & 43.1\% \\
  bernoulli                        & 88.2\% & 28.6\% \\
  binary\_cross\_entropy\_with\_logits & 84.5\% & 28.5\% \\
  cumsum                           & 86.6\% & 43.7\% \\
  div                              & 67.5\% & 26.4\% \\
  empty\_strided                    & 87.0\% & 52.8\% \\
  huber\_loss                       & 96.0\% & 25.1\% \\
  i0                               & 90.6\% & 16.7\% \\
  linear                           & 91.4\% & 58.7\% \\
  matmul                           & 92.2\% & 34.8\% \\
  mean                             & 97.1\% & 28.9\% \\
  mish\_backward                    & 94.7\% & 43.6\% \\
  mm                               & 94.4\% & 58.2\% \\
  mul                              & 97.0\% & 30.7\% \\
  poisson                          & 87.3\% & 21.3\% \\
  pow                              & 92.5\% & 36.0\% \\
  rrelu\_with\_noise                 & 91.0\% & 45.0\% \\
  sort                             & 71.2\% & 17.8\% \\
  sub                              & 91.2\% & 42.3\% \\
  \midrule
  \textbf{Overall Avg.}            & \textbf{88.4\%} & \textbf{36.3\%} \\
  \bottomrule
  \end{tabular}
  }
\end{minipage}\hfill
\begin{minipage}[c]{0.53\textwidth}
  \centering
  \includegraphics[width=\textwidth]{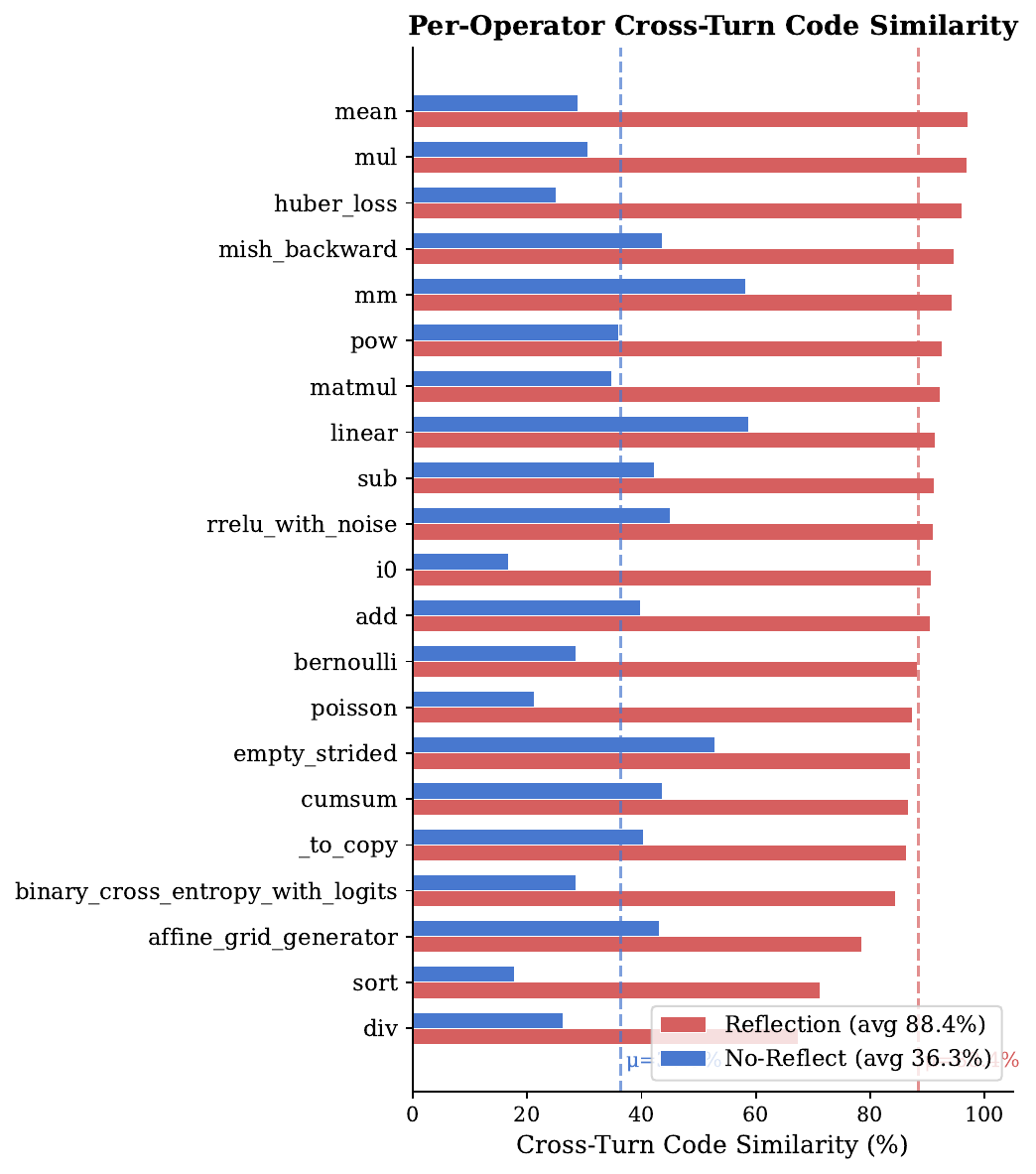}
  \caption{\small Cross-turn similarity for 21 jointly failed operators. Reflection clusters high (local patching); No-Reflect spreads wide (global search).}
  \label{fig:reflection_detail}
\end{minipage}
\end{figure}

Table~\ref{tab:reflection_similarity} compares the cross-round code similarity for 21 operators that both Reflection and No-Reflection modes fail to solve. Similarity is computed via line-level comparison using \texttt{difflib.SequenceMatcher}, with each operator averaged over 4 round transitions (round $i \to i{+}1$).

\clearpage

\end{document}